\documentclass[11pt]{article}

\usepackage[preprint]{acl}

\usepackage{times}
\usepackage{latexsym}

\usepackage[T1]{fontenc}

\usepackage[utf8]{inputenc}

\usepackage{microtype}

\usepackage{inconsolata}

\usepackage{graphicx}

\usepackage{xspace}
\usepackage{multirow}
\usepackage{booktabs}
\usepackage{enumitem}
\usepackage{tcolorbox}
\usepackage{pifont}
\usepackage[table]{xcolor}
\usepackage{CJKutf8}
\usepackage{amsmath}
\usepackage{enumitem}
\usepackage{listings}
\usepackage{hyperref}
\usepackage{hhline}
\usepackage{authblk}

\newcommand{\realtask}{appellate review\xspace}
\newcommand{\task}{\textsc{appellate review}\xspace}
\newcommand{\taskbold}{\textbf{\task}\xspace}

\newcommand{\dataset}{\textsc{AR-Bench}\xspace}
\newcommand{\datasetbold}{\textbf{\dataset}\xspace}

\newcommand{\cmark}{\ding{51}}
\newcommand{\xmark}{\ding{55}}

\newcommand{\pmtDetection}{Prompt for Error Detection.}
\newcommand{\pmtClassification}{Prompt for Error Classification.}
\newcommand{\pmtChargeCorrection}{Prompt for Charge Error Correction.}
\newcommand{\pmtTermCorrection}{Prompt for Prison Term Error Correction.}
\newcommand{\pmtFineCorrection}{Prompt for Fine Error Correction.}

\definecolor{highlightcolor}{HTML}{FFBF31}

\urldef{\myurl}\url{http://gongbao.court.gov.cn/Details/415f49dd8baaa04b479d57af9616ef.html?sw=%e4%b8%ad%e5%9b%bd%e8%a3%81%e5%88%a4%e6%96%87%e4%b9%a6}

\title{\dataset: Benchmarking Legal Reasoning with Judgment Error Detection, Classification and Correction}

\author{
    \textbf{Yifei Li\textsuperscript{1}},
    \textbf{Richong Zhang\textsuperscript{1}}\thanks{Corresponding author},
    \textbf{Wanyu Tu\textsuperscript{2}},
    \textbf{Zhijie Nie\textsuperscript{1}},
    \textbf{Haokun Luo\textsuperscript{1}},
    \\
    \textbf{Chuantao Yin\textsuperscript{3}},
    \textbf{Pengchong Li}\textsuperscript{4}
    \\
    \textsuperscript{1}SKLCCSE, Beihang University, China,
    \\
    \textsuperscript{2}SCCE, University of Science and Technology Beijing, China,
    \\
    \textsuperscript{3}Sino-French Engineer School, Beihang University, China,
    \\
    \textsuperscript{4}People's Procuratorate of Beijing Municipality, China
    \\
    \texttt{\{liyifei,zhangrc\}@act.buaa.edu.cn}
}

\begin{document}
\maketitle

\begin{abstract}
Legal judgments may contain errors due to the complexity of case circumstances and the abstract nature of legal concepts, while existing appellate review mechanisms face efficiency pressures from a surge in case volumes. Although current legal AI research focuses on tasks like judgment prediction and legal document generation, the task of judgment review differs fundamentally in its objectives and paradigm: it centers on detecting, classifying, and correcting errors after a judgment is issued, constituting anomaly detection rather than prediction or generation. To address this research gap, we introduce a novel task \taskbold, aiming to assess models' diagnostic reasoning and reliability in legal practice. We also construct a novel dataset benchmark \datasetbold, which comprises 8,700 finely annotated decisions and 34,617 supplementary corpora. By evaluating 14 large language models, we reveal critical limitations in existing models' ability to identify legal application errors, providing empirical evidence for future improvements.
\end{abstract}
\section{Introduction}

\begin{CJK}{UTF8}{gkai}
刑当罪则威，不当罪则侮。
\hfill \mbox{--- 荀子}
\vspace{0.2 em}

\noindent \textit{A fitting penalty commands authority, an unjust one incurs contempt.} 
\hfill \mbox{--- \textit{Xunzi [c. 310--c. 235 BC]}}
\end{CJK}
\vspace{0.5 em}

Equality before the law is a fundamental principle of modern justice, crucial for ensuring balanced sentencing in criminal law. Although judicial rulings undergo rigorous procedures, judges may still make errors during the adjudication process due to the complexity of cases and the abstract nature of legal norms~\cite{Hu2024}. Both civil law~\citep{Hoquet1981, Woo1989, Lindemann2020} and common law~\citep{Orfield1936, Hessick2008} systems have established appeal or review mechanisms to reexamine the accuracy of legal application and the appropriateness of sentencing, thereby correcting potential errors.

Taking China as an example, the public prosecution service (or procuratorate) may review finalized judgments regarding issues such as the correctness of legal application and, where appropriate, file a protest or request for retrial in accordance with the law. These \realtask mechanisms help enhance the accuracy and fairness of judgments, correct errors in the application of law~\citep{Bai2006}, and curb potential abuses of judicial discretion.

\begin{figure}[t]
  \centering
  \includegraphics[width=\columnwidth]{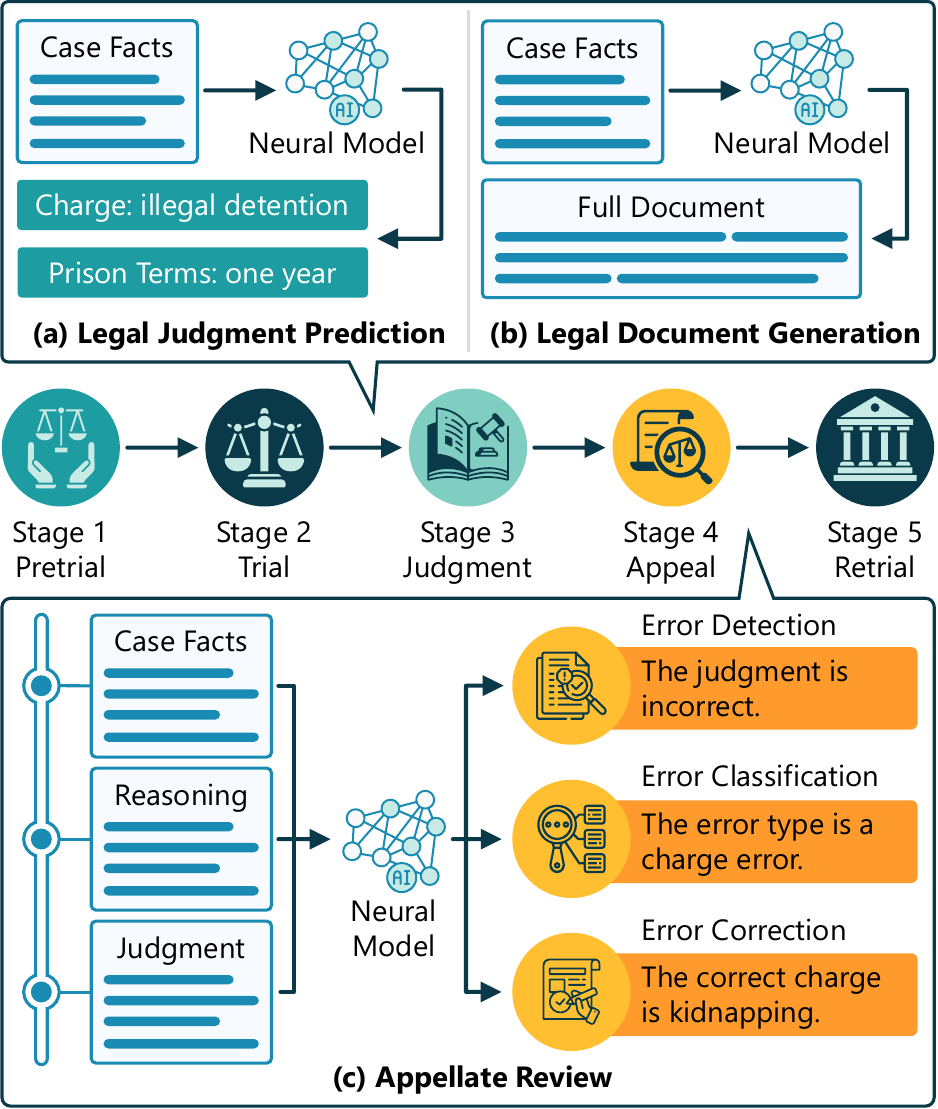}
  \caption{Difference between \textbf{(a) legal judgment prediction}, \textbf{(b) legal document generation}, and \textbf{(c) \realtask}.}
  \label{fig:intro}
\end{figure}

However, this \realtask mechanism now faces practical pressures~\citep{Wallace2005, Zuo2018}. In China, the number of appeal cases increased by 65.8\% from 2015 to 2024~\citep{Gongbao:JudicialStatsChina}. With the rapid increase in the number of caseloads~\cite{Lavie2016} and law articles~\citep{Tate1972}, judges' workloads have grown increasingly heavy, which may diminish the thoroughness of review and the effectiveness of error correction. Against this backdrop, a critical question arises: \textit{Can artificial intelligence (AI) be introduced to assist in the review of finalized judgments?}

While existing research in legal AI has predominantly focused on tasks such as legal judgment prediction~\cite{Cail2018, Chalkidis2019} and legal document generation~\cite{CaseSummarizer2016, Shen2022}, Figure \ref{fig:intro} highlights two primary distinctions between them and \realtask. 

First, they target fundamentally different stages of the legal process. Tasks such as legal judgment prediction~\citep{Feng2022} and legal document generation~\citep{Su2025} focus on the period before a judgment is rendered, aiming to assist legal practitioners in predicting outcomes or generating documents. In contrast, \realtask focus on the period after a judgment is rendered, aiming to examine existing judgments and provide references for determining whether to initiate a retrial. 

A further distinction lies in their fundamentally different learning objectives. Legal judgment prediction involves predicting legal outcomes from fact description, constituting a classification or prediction task; \realtask identifies potential errors within finalized judgments, falling under error identification or anomaly detection tasks~\citep{Felice2015, Yang2025}. Similarly, legal document generation produces structured text from input information, akin to conditional text generation whereas \realtask deconstructs and diagnoses existing texts, aligning more closely with fact-checking~\citep{FEVER2018, Schlichtkrull2023} paradigms.

These distinctions implies that \realtask cannot be encompassed within existing tasks. However, this critical question has yet to be systematically explored: the task of specifically reviewing judicial errors in legal judgment still lacks a standardized evaluation benchmark.

\begin{table*}[t]
\centering
\resizebox{\textwidth}{!}
{
\begin{tabular}{l|cccccccc}
\toprule
\textbf{Benchmark} & 
\textbf{\begin{tabular}[c]{@{}c@{}}
Legal\\Facts\end{tabular}}& 
\textbf{\begin{tabular}[c]{@{}c@{}}
Reasoning\\Process\end{tabular}} & 
\textbf{\begin{tabular}[c]{@{}c@{}}
Law\\Articles\end{tabular}} &
\textbf{\begin{tabular}[c]{@{}c@{}}
Sentencing\\Factors\end{tabular}} & 
\textbf{\begin{tabular}[c]{@{}c@{}}
Label\\Classes\end{tabular}} & 
\textbf{\begin{tabular}[c]{@{}c@{}}
Task\\Classes\end{tabular}} & 
\textbf{Type of Cases} & 
\textbf{Jurisdiction} \\
\midrule
\textbf{CAIL2018~\citep{Cail2018}} 
& \cmark & \xmark & \xmark & \xmark & 2 & 2 
& Criminal & Mainland China \\
\textbf{Auto-Judge~\cite{Long2019}} 
& \cmark & \xmark & \xmark & \xmark & 2 & 1 
& Civil & Mainland China \\
\textbf{SwissJP~\cite{SwissBenchmark2021}}
& \cmark & \xmark & \xmark & \xmark & 2 & 1 
& Generic & Switzerland \\
\textbf{LBOX OPEN~\cite{Hwang2022}} 
& \cmark & \xmark & \xmark & \xmark & 4 & 2 
& Generic & Korean \\
\textbf{JuDGE~\cite{Su2025}} 
& \cmark & \cmark & \cmark & \xmark & 8 & 1
& Criminal & Mainland China \\
\textbf{\datasetbold (Ours)} 
& \cmark & \cmark & \cmark & \cmark & 20 & 3
& Criminal & Mainland China \\
\bottomrule
\end{tabular}
}
\caption{Comparison between \dataset and existing legal judgment prediction benchmarks in civil law systems.}
\label{tab:ar_compare}
\end{table*}

To address this research gap, we propose a novel task \taskbold, designed to specifically evaluate models' ability to identify errors in legal application within legal judgment. We decompose \task into three sequential subtasks: error detection, error classification, and error correction, providing a clear framework for systematic evaluation. Given an initial case fact description and a corresponding judgment, models are tasked to: (1) detect whether the judgment contains legal errors; (2) identify the specific type(s) of error; (3) generate a corrected, legally valid revision based on the identified error.

To support this systematic evaluation framework, we construct a novel benchmark, \datasetbold, based on real-world cases. The benchmark comprises 8,700 finely annotated judgments and a supplementary corpus of 34,617 judgments, providing \textit{20 fine-grained annotations}, including resoning process, law articles, and sentencing factors. It supports \textit{3 progressive subtasks}: error detection, classification, and correction, which significantly surpasses prior work in both annotation granularity and task complexity. We conduct a thorough evaluation of 14 large language models (LLMs), including general-domain, reasoning-enhanced, and domain-specific LLMs, revealing their critical limitations in handling complex legal logic and providing empirical evidence for potential improvements.

We summarize our contributions as follows:
\begin{itemize}[leftmargin=*]
\item We propose \task, a novel task designed to evaluate LLMs' core capability in detecting, classifying, and correcting errors in legal judgments. Unlike prior tasks focused legal judgment prediction, legal case retrieval or legal document generation, our task directly assesses models' diagnostic reasoning in legal practice.
\item To address the abstractness of legal concepts and the complexity of case scenarios, we innovatively decompose the task into three progressive subtasks: error detection, error classification, and error correction. This provides a structured framework for systematic evaluation.
\item We construct \dataset, a large-scale benchmark dataset and conduct comprehensive evaluations of mainstream LLMs. Our results not only reveal key limitations in handling complex legal logic but also provide empirical foundations for future improvements in legal AI.
\end{itemize}
\section{Related Work}
\subsection{Legal Judgment Prediction}
Legal judgment prediction has undergone a technological evolution from rule-based/statistical methods~\citep{Lawlor1963, Ulmer1963} and traditional machine learning~\cite{Liu2006, Aletras2016} to deep learning and pre-trained models~\citep{Feng2022, Wu2023, Zhang2023, Tyss2024, Yue2024}, with its research objective remaining consistent: predicting legal outcomes based on case fact descriptions. Although judgment prediction models  have made some progress, they still face two core challenges: the precise differentiation of highly detail-sensitive, easily confused criminal charges~\citep{Xu2020}, and the accurate prediction of numerical information such as prison terms and involved amounts~\citep{Gan2023}. In response, a range of methods are proposed, such as enhancing models' sensitivity to subtle factual distinctions through contrastive learning~\citep{Liu2022, Zhang2023} methods and modeling legal knowledge using graph structures~\citep{Bi2023, Li2024}. However, these efforts remain focused on the prediction paradigm.

\subsection{Legal Reasoning Benchmarks} 
Unlike judgment prediction focusing on predicting judgments based on fact descriptions, \task aims to identify potential errors in final judgments from finalized judgment documents. This requires models to possess deeper legal logical analysis and critical reasoning capabilities. Current judgment prediction datasets~\citep{Cail2018, Hu2018, Chalkidis2019} are primarily designed for fine-tuning small language models (SLMs) and lack fine-grained annotations. With the rise of large language models, a series of benchmarks have emerged to evaluate their legal reasoning capabilities, covering both comprehensive assessments~\citep{Guha2023, Fei2024, LexEval2024} and specialized task~\citep{Yuan2024, Niklaus2025, LegalAgentBench2025} evaluations. However, their task definitions remain largely confined to existing tasks such as legal judgment prediction~\citep{An2022, Shui2023, LJPCheck2024}, legal document retrieval~\citep{Upadhya2025, Seo2025}, and legal document generation~\citep{Shen2022, Su2025, Rolshoven2025}. They have yet to encompass \realtask—a critical thinking paradigm that is both essential to judicial practice and highly challenging, as summarized in Table~\ref{tab:ar_compare}.
\section{\dataset}
We introduce \dataset, a novel benchmark designed to systematically evaluate LLMs' diagnostic reasoning and reliability in legal practice through three sequential tasks: error detection, error classification, and error correction. \dataset aims to assess whether current LLMs can assist judges in reviewing finalized judgments.

\subsection{Task Design and Formulation}
\label{sec:task_formulation}

\begin{figure*}[t]
  \centering
  \includegraphics[width=\textwidth]{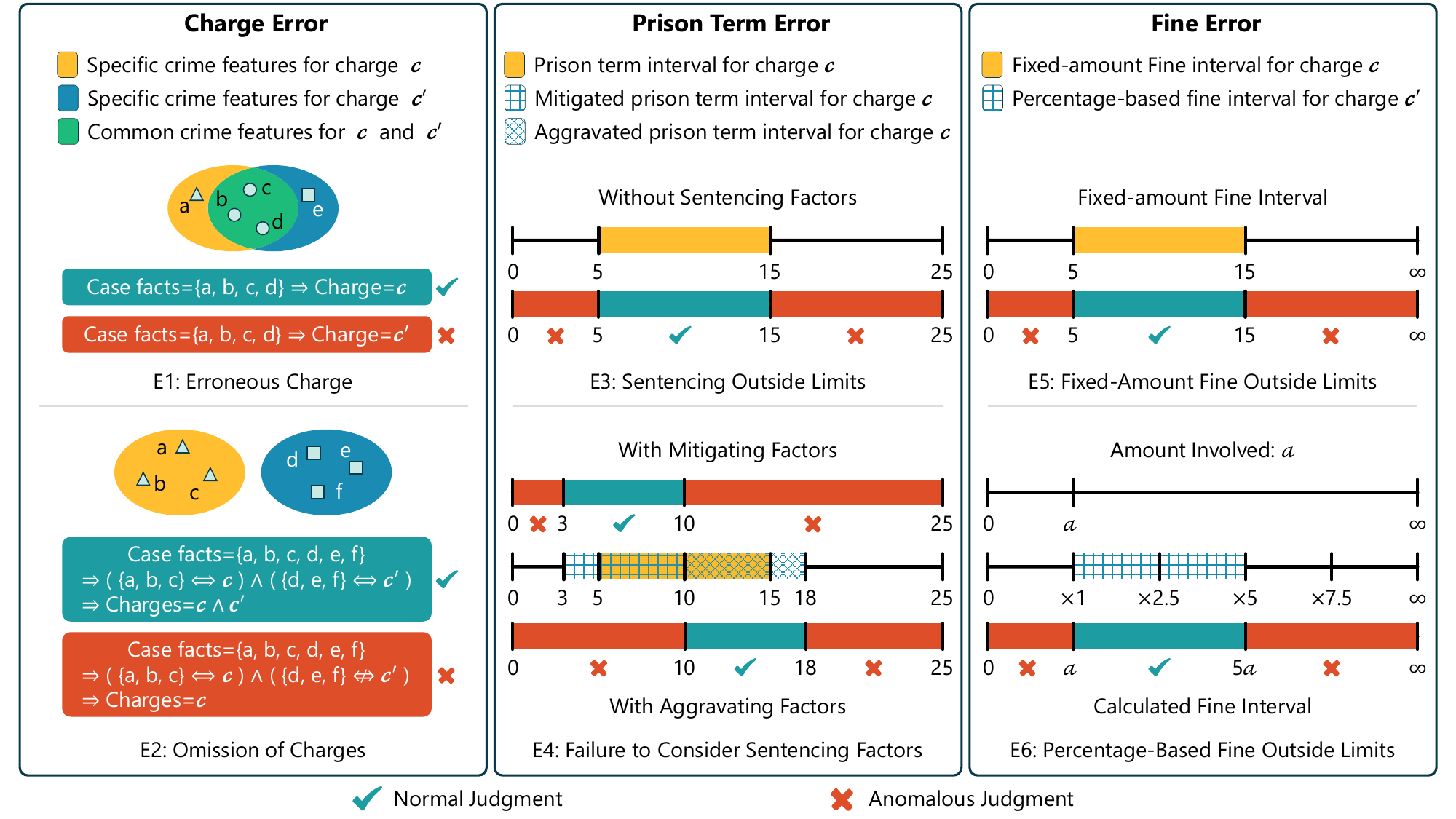}
  \caption{Taxonomy of Six Manually Constructed Error Types for \dataset. This diagram illustrates the designed errors targeting three adjudication elements: \textbf{Charges} (E1: Erroneous Charge, E2: Omission of Charges), \textbf{Prison Terms} (E3: Sentencing Outside Limits, E4: Failure to Consider Sentencing Factors), and \textbf{Fines} (E5: Fixed-amount Fine Outside Limits, E6: Percentage-based Fine Outside Limits).}
  \label{fig:error}
\end{figure*}

\dataset evaluates models across three integral tasks, each targeting specific dimensions of legal reasoning and knowledge.  This enables comprehensive model performance evaluation, covering the entire process from error detection to correction, as illustrated in Figure \ref{fig:intro}.

\vspace{0.5 em}
\textbf{Error Detection.} In the process of \realtask, judges must first determine whether the trial court's judgment contains errors, and then conduct a detailed examination of the erroneous judgment. This task tests the model's ability to identify the presence of judicial errors of law application within a given judgment document $\mathcal{D} $. The model is expected to output a binary answer $\hat{y} \in \{ \text{true}, \text{false}\}$, where $\text{true}$ indicates the judgment is correct, and $\text{false}$ indicates the judgment contains error. 

\vspace{0.5 em}
\textbf{Error Classification.} Once it is determined that an error exists in the judgment, the next step is to focus on what type of error is present in the judgment. This task assesses whether the model can classify the type of specific errors in the given judgment document $\mathcal{D}$. The model is expected to output a predicted error type $\hat{E}$ that it believes are present, chosen from three predefined error categories $\mathcal{E}$: charge, prison term, and fine, as shown in Figure \ref{fig:error}. 
We consider that multiple types of errors may occur simultaneously. However, considering the difficulty of evaluating multiple corrections in the subsequent task, we restrict the model to predicting the primary error in the judgment.

\vspace{0.5 em}
\textbf{Error Correction.} To assess whether the model has gained a deep understanding of the root causes of errors, we designed error correction to further evaluate its legal reasoning capabilities. Error correction evaluates the model's capacity to generate accurate local corrections for a given error type $E$ within a judgment document $\mathcal{D}$. The model is expected to output a local correction $\hat{C}$, as an alternative of the original erroneous judgment.

\subsection{Evaluation Metric}
For error detection and error classification tasks, we employ classification metrics including Accuracy (Acc), Macro-Precision (MaP), Macro-Recall (MaR), and Macro-F1 (MaF1) to better evaluate model performance under category imbalance.
For the error correction, we employ separate evaluation metrics for different error types. The aforementioned classification metrics are used for corrections pertaining to criminal charges, whereas the classification metrics, ImpScore and Acc@0.1~\citep{Chen2019} are employed to evaluate those addressing prison terms and fines.

\subsection{Data Collection and Preparation}
We introduce the data collection and preparation process of \dataset, covering initial acquisition through to data preprocessing.

\begin{figure}[t]
  \centering
  \includegraphics[width=\columnwidth]{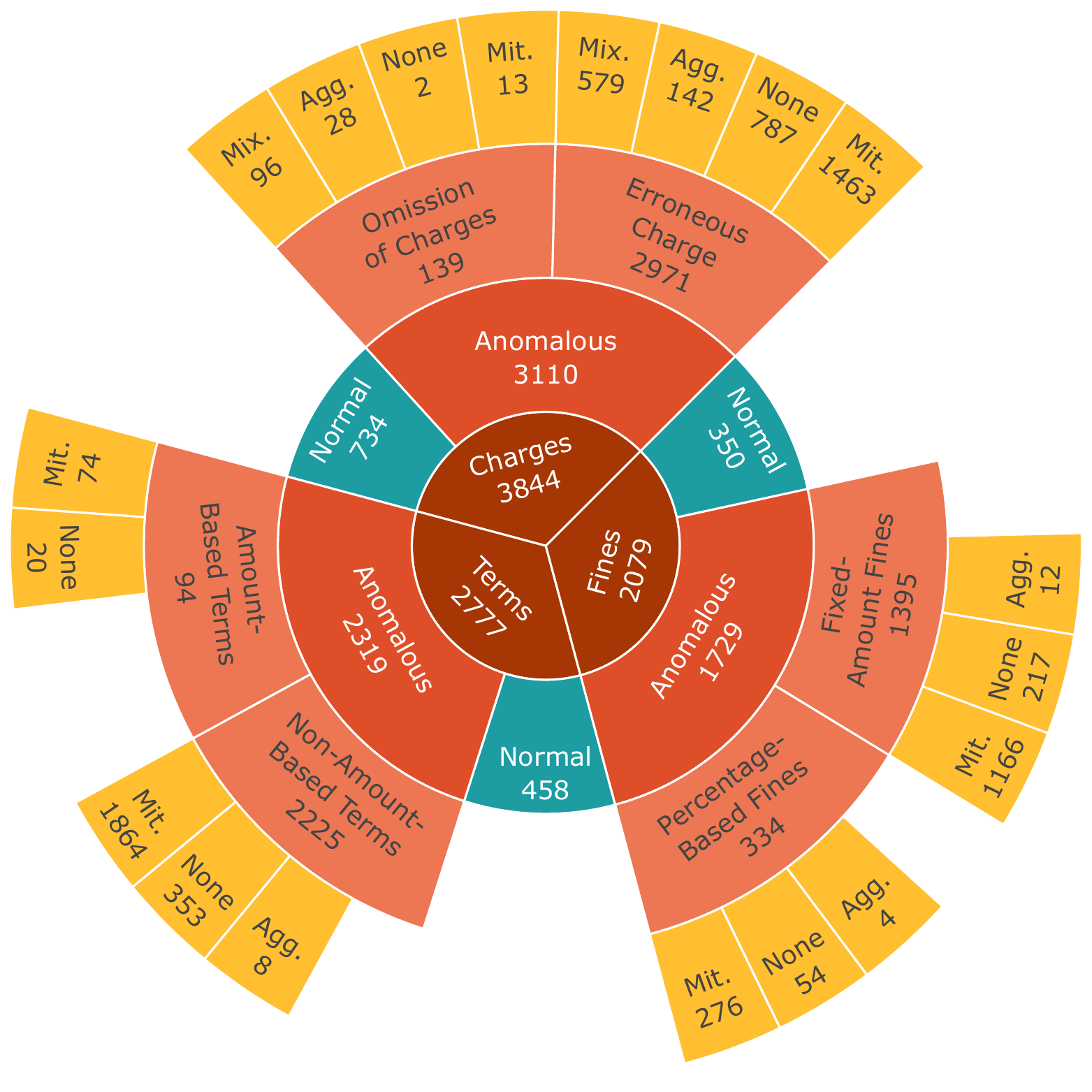}
  \caption{Distribution of \dataset, highlighting normal versus anomalous cases and further breaking anomalies down by error type (e.g., omission, erroneous charges, fixed amounts, percentage-based fines) and sentencing factors (None, Mitigated, Aggravated, Mixed).
  }
  \label{fig:chart}
\end{figure}

\vspace{0.5 em}
\textbf{Data Acquisition.} \dataset is derived from an open-source legal document generation dataset\footnote{\url{https://github.com/oneal2000/JuDGE}} \citep{Su2025}, with raw data obtained from China Judgments Online\footnote{\url{https://wenshu.court.gov.cn/}}. Although it includes preliminary annotations, initial analysis has identified several quality issues: (1) Incomplete annotations. The reasoning process lacks comprehensive annotations of sentencing factors. (2) Annotation errors. Some judgment samples contain inaccurately extracted charge labels, especially in multi-charge cases where historical cumulative charges are often conflated with the charges currently under adjudication. (3) Sample duplication. The original dataset includes redundant entries. Therefore, we manually cleaned and relabeled the data to construct error judgment samples for subsequent analysis. Detailed data license can be found in Appendix~\ref{app:license}.

\vspace{0.5 em}
\textbf{Data Processing.} 
First, we merged the original dataset with an external corpus of judgment documents and removed duplicate samples. Second, due to the complexity of reviewing judgments in multi-defendant cases, we excluded all such cases, retaining only those involving a single defendant. Third, we used regular expressions to re-extract the fields requiring annotation and mapped them to a unified format. Finally, to evaluate the model' s reasoning capability given external legal knowledge , we compiled the full text of the legal articles corresponding to all cited article numbers.

\subsection{Judgment Errors Creation}
We manually alter the judgment document to introduce errors, resulting in reasoning and judgments that are anomalous and inconsistent with the reference judgment. Our dataset includes 6 error scenarios relevant to real-world \realtask applications, as shown in Figure~\ref{fig:error}, with detailed annotation process can be found in Appendix~\ref{app:annotation}.

\vspace{0.5 em}
\textbf{Charge Error.} 
Charge Errors fall into two types: Erroneous Conviction (E1) and Omission of Charges (E2). 
Erroneous conviction (E1) refers to situations where two charges are easily confused in a judgment~\cite{Bai2006} due to highly similar criminal characteristics. 
Omission of charges (E2) refers to situations where a case should involve multiple offenses, but the initial judgment only recognized the primary offense corresponding to the main criminal characteristics while overlooking other offenses matching different characteristics. 

\begin{table}[!t]
\centering
\resizebox{\columnwidth}{!}{%
\begin{tabular}{lccc}
\toprule
Component & Max Length & Min Length & Mean Length \\
\midrule
Fact Description & 1692.0 & 89.0 & 380.9\\
Reasoning Process & 2567.0 & 97.0 & 286.5\\
Judgment & 1027.0 & 19.0 & 87.6\\
Law Article & 1622.0 & 39.0 & 557.9\\
\bottomrule
\end{tabular}%
}
\caption{Character-level length distribution of sample components in our \dataset.}
\label{tab:stats_length}
\end{table}

\vspace{0.5 em}
\textbf{Prison Term Error.} 
Prison Term (or Sentencing) errors fall into two categories: Sentencing Outside Limits (E3) and Failure to Consider Sentencing Factors (E4). 
Sentencing outside limits (E3) refers to sentences imposed that exceed the statutory sentencing limits specified in criminal law. 
Failure to consider sentencing factors (E4) refers to a sentencing determination that deviates from the sentencing range established based on such factors, as illustrated in the lower half of Figure~\ref{fig:error}. 

\vspace{0.5 em}
\textbf{Fine Error.} Fines errors encompass both Fixed-Amount Fine Outside Limits (E5) and Percentage-based Fine Outside Limits (E6). Fixed-amount fine outside limits (E5) occurs when the Criminal Law specifies a fixed fine range (for 5 to 15 as shown in Figure~\ref{fig:error}), and a judge imposes a fine exceeding this range. Percentage-based fine outside limits (E6) occurs when the Criminal Law establishes a fine range in proportional terms (from $\times 1 $ to $\times 5$ as shown in Figue~\ref{fig:error}), and a judge fails to accurately determine the involved amount during sentencing, resulting in a judgment error where the imposed fine exceeds the proportional range.

\subsection{Dataset Statistics}

Figure~\ref{fig:chart} displays the statistical counts of different sample types in \dataset using a fine-grained taxonomy, because distinct annotation workflows are required during data construction based on different category. For instance, errors in charge determination require matching the most similar cases based on sentencing factors, while prison term and fine errors necessitate determining the starting point based on whether amounts are involved. The data distribution is uneven, consistent with real-world data patterns and widely seen in prior works~\citep{Hu2018, Zhang2023}. 

Table~\ref{tab:stats_length} also presents the text length distribution of \dataset. Notably, law articles are significantly longer than other judgment content, making the effective utilization of knowledge embedded in the original legal text a key challenge.

\begin{table*}[t]
\centering
\resizebox{\textwidth}{!}{%
\begin{tabular}{lcccccccccccc}
\toprule
 \multirow{2}{*}{Models} & \multicolumn{4}{c}{Error Detection} & \multicolumn{4}{c}{Error Classification}                          & \multicolumn{4}{c}{Error Correction}                   \\ \cmidrule(lr){2-5} \cmidrule(lr){6-9} \cmidrule(lr){10-13}
& Acc & MaP & MaR & MaF1         
& Acc & MaP & MaR & MaF1 
& Acc & MaP & MaR & MaF1 \\
 \midrule
 \rowcolor{highlightcolor!50} 
 \multicolumn{13}{c}{\textsc{General-domain LLMs}}   \\
GPT-5.1  & 86.40 & 78.63 & 85.67 & 81.14 & 89.36 & 89.32 & 89.35 & 89.08  & 50.86 & 48.09 & 45.36 & 44.68\\ 
Qwen3-Max  & \underline{92.03} & 86.57 & \textbf{89.78} & \textbf{88.03} & \textbf{94.75} & \textbf{95.01} & \textbf{94.05} & \textbf{94.48} & \textbf{74.10} & \underline{67.75} & 64.26 & \underline{64.42}\\ 
DeepSeek-V3.2-Chat & 78.87 & 71.54 & 80.47 & 73.26
 & 84.80 & 86.71 & 82.19 & 83.96  & 65.67 & 61.23 & 57.69 & 57.25\\ 
GLM-4.6-Chat  & 87.74 & 80.26 & 86.69 & 82.70 & 91.89 & 92.45 & 90.42 & 91.28 & 70.91 & 66.11 & 59.52 & 60.49\\ 
MiniMax-M2  & 72.47 & 67.65 & 76.74 & 67.53 & 88.20 & 86.39 & 87.11 & 86.68 & 57.54 & 58.42 & 50.97 & 51.16 \\ 
Qwen3-14B-Chat  & 88.05 & 80.66 & 87.25 & 83.16 & 85.41 & 87.72 & 84.55 & 84.78 & 53.06 & 52.67 & 41.82 & 41.35\\ 
Qwen3-8B-Chat  & 65.84 & 67.15 & 76.86 & 62.94 & 71.14 & 76.75 & 62.77 & 63.86 & 55.67 & 54.14 & 47.54 & 44.48 \\ 

\midrule
\rowcolor{highlightcolor!50} 
\multicolumn{13}{c}{\textsc{Reasoning-enhanced LLMs}} \\
GPT-4o  & 81.05 & 74.73 & 85.99 & 76.61 & 89.52 & 88.91 & 90.63 & 89.59 & 61.94 & 57.38 & 50.20 & 50.87\\ 
Qwen3-Max-preview-Thinking  & 91.84 & 87.65 & 86.25
 & \underline{86.93} & 91.91 & 91.12 & 90.83 & 90.95 & 70.88 & 66.64 & \underline{65.20} & 64.39\\ 
DeepSeek-V3.2-Thinking  & 87.53 & 80.17 & \underline{88.81} & 83.05 & 91.78 & 90.28 & 90.98 & 90.61 & \underline{72.61} & \textbf{69.60} & \textbf{67.14} & \textbf{66.56}\\ 
Kimi-K2-Thinking  & \textbf{92.29} & \textbf{90.38} & 84.20 & 86.82 & \underline{93.86} & \underline{93.31} & \underline{93.11} & \underline{93.21} & 72.60 & 65.68 & 63.55 & 63.05\\ 
Qwen3-Next-80B-A3B-Thinking  & 89.25 & \underline{88.28} & 75.95 & 80.08 & 88.74 & 88.02 & 86.72 & 87.31 & 63.80 & 57.98 & 57.05 & 55.41\\ 
\midrule
\rowcolor{highlightcolor!50} 
\multicolumn{13}{c}{\textsc{Domain-specific LLMs}} \\ 

Farui-Plus  & 48.12 & 60.23 & 64.00 & 47.34 & 66.38 & 63.43 & 64.33 & 63.72 & 52.65 & 53.34 & 42.16 & 41.33\\ 
LawLLM-7B  & 26.34 & 59.55 & 53.76 & 25.04 & 39.85 & 53.13 & 42.70 & 34.20 & 31.33 & 47.44 & 30.33 & 31.21\\ 
\bottomrule
\end{tabular}%
}
\caption{Performance comparison of large language models on error detection, error classification, and error correction tasks, reported in terms of Accuracy (Acc), Macro-Precision (MaP), Macro-Recall (MaR), and Macro-F1 (MaF1) across general-domain, reasoning-enhanced, and domain-specific LLMs.}
\label{tab:main}
\end{table*}

\section{Experiments}
The primary objective of this section is to systematically analyze current models' performance and limitations. We conduct experiment to address the following questions:
\begin{enumerate}[label=\textbf{RQ\arabic*:}, leftmargin=*, wide, nosep]
    \item How do existing LLMs perform on the three sequential tasks: error detection, error classification, and error correction?
    \item How do different components in judgment documents and additional knowledge (such as legal articles) affect model performance?
    \item Can legal judgment prediction models be applied to the error correction sub-task? How do they perform on existing dataset and \dataset? 
    \item How do existing models perform on \task across different error types?
    \item How do human perform on three sub-tasks of \task?
\end{enumerate}

\subsection{Baselines}
We categorize baselines into four groups: general-domain LLMs, reasoning-enhanced LLMs, domain-specific LLMs, and fine-tuned SLMs.

\vspace{0.5 em}
\textbf{General-Domain LLMs.} This group consists of 7 LLMs, including:
\texttt{GPT-5.1-2025-11-13}~\citep{GPT51}, 
\texttt{Qwen3-Max-2025-09-23}~\citep{Qwen3}, 
\texttt{DeepSeek-V3.2-Chat}~\citep{DeepSeekv32}, \texttt{GLM-4.6-Chat}~\citep{Glm2024}, 
\texttt{MiniMax-M2}~\citep{Minimax-m2}, 
\texttt{Qwen3-14B-Chat}~\citep{Qwen3}, and
\texttt{Qwen3-8B-Chat}~\citep{Qwen3}. 
These models are pre-trained on diverse, large-scale corpora to develop robust language understanding and generation capabilities.

\begin{table*}[t]
\centering
\resizebox{0.9\textwidth}{!}{
\begin{tabular}{l|cccccccccc}
 \toprule
 \multirow{2}{*}{Models} & \multirow{2}{*}{Training Set} &
 \multirow{2}{*}{Test Set}
 & \multicolumn{4}{c}{Charges}  & \multicolumn{4}{c}{Prison Terms} \\
\hhline{~|~~|-|-|-|-|-|-|-|-|}
 & & 
   & Acc & MaP & MaR & MaF1 
   & Acc & MaP & MaR & MaF1 \\
 \midrule
\multirow{4}{*}{TopJudge} 
& \multirow{2}{*}{CAIL-S} & \textsc{CAIL-S} & 71.61 & 66.83 & 73.43 & 67.28 & 34.65 & 27.19 & 30.91 & 26.47 \\
&   & \dataset & 90.32 & 46.83 & 45.73 & 43.41 & 62.30 & 18.70 & 23.27 & 18.20 \\
\hhline{~|-|-|-|-|-|-|-|-|-|-|}
& \multirow{2}{*}{CAIL-B} & \textsc{CAIL-B} & 85.49 & 33.40 & 36.60 & 32.18 & 49.77 & 26.24 & 29.62 & 26.57\\
&  & \dataset & 91.60 & 26.75 & 36.54 & 29.64 & 63.67 & 20.57 & 20.40 & 18.19 \\
\hline
\multirow{4}{*}{LADAN} 
& \multirow{2}{*}{CAIL-S} & \textsc{CAIL-S} & 84.67 & 82.58 & 82.35 & 81.93 & 35.57 & 31.25 & 32.58 & 31.76 \\
&   & \dataset & 93.72 & 55.75 & 51.65 & 52.36 & 63.33 & 23.90 & 26.07 & 23.44 \\
\hhline{~|-|-|-|-|-|-|-|-|-|-|}
& \multirow{2}{*}{CAIL-B} & \textsc{CAIL-B} & 96.01 & 82.00 & 86.05 & 83.36 & 58.77 & 43.60 & 49.63 & 45.19 \\
&  & \dataset & 95.66 & 49.38 & 57.95 & 52.89 & 67.45 & 33.49 & 37.76 & 33.46 \\
\hline
\multirow{4}{*}{NeurJudge} 
& \multirow{2}{*}{CAIL-S} & \textsc{CAIL-S} & 83.06 & 81.04 & 81.41 & 80.60 & 37.79 & 32.11 & 35.79 & 33.56\\
&   & \dataset & 90.23 & 85.41 & 84.43 & 84.30 & 64.34 & 35.57 & 30.58 & 28.23 \\
\hhline{~|-|-|-|-|-|-|-|-|-|-|}
& \multirow{2}{*}{CAIL-B} & \textsc{CAIL-B} & 89.07 & 55.08 & 71.44 & 58.41 & 50.58 & 24.52 & 39.60 & 26.39 \\
&  & \dataset & 89.56 & 79.78 & 85.44 & 81.25 & 63.14 & 22.51 & 28.21 & 21.59 \\
\bottomrule
\end{tabular}%
}
\caption{Performance comparison of TopJudge, LADAN, and NeurJudge trained on CAIL small (CAIL-S) and CAIL big (CAIL-B), evaluated on CAIL and \dataset test sets, reporting Accuracy (Acc), Macro Precision (MaP), Macro Recall (MaR), and Macro F1 (MaF1) for charge prediction and prison term prediction. }
\label{tab:slm}
\end{table*}

\vspace{0.5 em}
\textbf{Reasoning-Enhanced LLMs.} The group of reasoning-enhanced LLMs contains a total of 5 LLMs, including: \texttt{GPT-4o}~\citep{GPT4o}, 
\texttt{Qwen3-Max-preview-Thinking}~\citep{Qwen3}, 
\texttt{DeepSeek-V3.2-Thinking}~\citep{DeepSeekv32}, 
\texttt{Kimi-K2-Thinking}~\citep{KimiK2}, 
\texttt{Qwen3-Next-80B-A3B-Thinking}~\citep{Qwen3}.
In contrast to general-domain LLMs that provide direct responses, these models are designed to generate internal Chain-of-Thought processes before arriving at a final answer. They have been optimized through Reinforcement Learning to produce extended reasoning paths to enhance the precision of complex outputs.

\vspace{0.5 em}
\textbf{Domain-Specific LLMs.} This group of LLMs is composed of \texttt{Farui-Plus}~\citep{Farui}, and \texttt{LawLLM-7B}~\citep{lawllm}. Domain-specific large models are typically fine-tuned based on general-domain LLMs, utilizing legal domain corpora and task-optimized objectives to enhance the model's capability in handling legal tasks.

\vspace{0.5 em}
\textbf{Fine-tuned SLMs.} This group of SLMs includes \texttt{TopJudge}~\citep{Zhong2018}, \texttt{LADAN}~\citep{Xu2020}, and \texttt{NeurJudge}~\citep{Yue2021}. These models are SOTA pre-trained models fine-tuned for legal judgment prediction. We use charge error correction to assess their generalization performance on our dataset.

\begin{table}[t]
\centering
\resizebox{\columnwidth}{!}{
\begin{tabular}{ccccccc}
 \toprule
 \multirow{2}{*}{Settings} 
 & \multicolumn{2}{c}{Error Detection}
 & \multicolumn{2}{c}{Error Classification}                    & \multicolumn{2}{c}{Error Correction} \\
\cmidrule(lr){2-3}\cmidrule(lr){4-5}\cmidrule(lr){6-7}
   & Acc & MaF1 
   & Acc & MaF1 
   & Acc & MaF1 \\
 \midrule
\textsc{S1} & 91.79 & 87.09 & 92.56 & 48.07 & 86.30 & 86.50 \\
\textsc{S2} & 97.06 & 95.09 & 99.97 & 49.99 & 97.71 & 96.01 \\
\textsc{S3} & 95.79 & 92.78 & 99.97 & 49.99 & 97.34 & 95.46\\
\textsc{S4} & 94.63 & 91.47 & 96.67 & 49.15 & 95.46 & 90.93 \\
\textsc{S5} & 92.79 & 88.29 & 94.51 & 48.59 & 94.14 & 88.59 \\

\bottomrule
\end{tabular}%
}
\caption{Performance comparison of Qwen3-Max across 5 settings (S1–S5) on error detection, error classification, and error correction tasks, reported in terms of accuracy (Acc) and macro F1-score (MaF1).}
\label{tab:settings}
\end{table}

\subsection{Input Settings}
We compare the impact of different input components on models' performance:
\begin{enumerate}[label=\textbf{S\arabic*:}, leftmargin=*, wide, nosep]
    \item \textit{Case facts + anomalous judgment} aligns with current legal judgment prediction paradigm.
    \item \textit{S1 + normal reasoning process} contains sentencing factors and cited law article numbers.
    \item \textit{S2 + relevant law articles} provides law articles as additional legal knowledge.
    \item \textit{Case facts + anomalous reasoning process + anomalous judgment} as the most realistic setting.
    \item \textit{S4 + erroneously cited law articles} provides anomalous legal knowledge.
\end{enumerate}

\begin{figure*}[t]
  \centering
  \includegraphics[width=\textwidth]{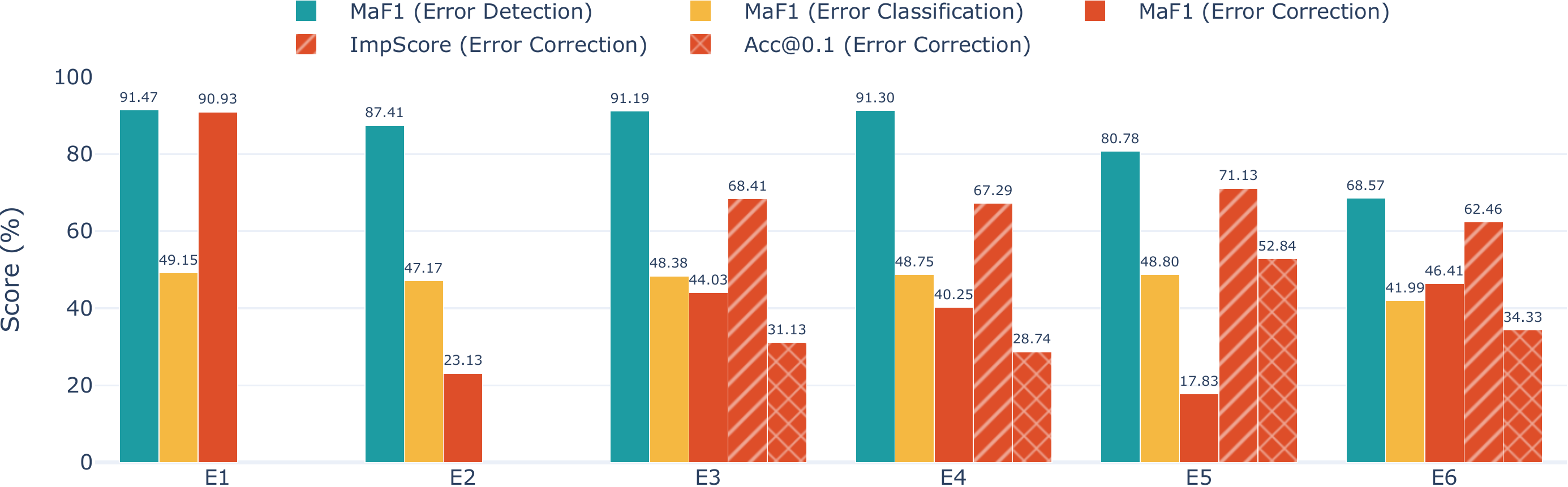}
  \caption{Performance comparison across different error types E1–E6 using error detection, classification, and correction metrics, including F1 scores, ImpScore, and correction accuracy at 0.1 tolerance.}
  \label{fig:exp_error_types}
\end{figure*}

\subsection{Main Results (RQ1)}
Table~\ref{tab:main} report the performance scores of all evaluated LLMs on three subtasks of \dataset. 

\textbf{Performance across tasks.} Error detection appears to be the most tractable task, error classification performs slightly better than detection on average, but models exhibit weaker instruction-following behavior, with further discussion in Appendix~\ref{app:instruction_following_rate}. In contrast, error correction is substantially more challenging for all models, with a clear performance drop across all metrics, reflecting the higher difficulty of generating correct revisions rather than detecting or classifying errors.

\textbf{Performance model families.} General-domain LLMs consistently achieve the strongest performance on error detection and classification, demonstrating robust legal understanding. Reasoning-enhanced models outperform general-domain models on the error correction task, suggesting that explicit or implicit reasoning mechanisms are beneficial when models must infer and generate corrected outputs. Domain-specific legal models lag significantly behind the other two categories across all tasks; notably, they underperform even when compared with general-domain models of similar parameter scales, indicating limitations that cannot be attributed solely to model size.

\subsection{Different Input Settings (RQ2)}
Based on the results in Table~\ref{tab:settings}, three key observations emerge. Introducing explicit reasoning substantially improves judgment review performance, as S2 and S4 consistently outperform the basic setting S1 across all sub-tasks. In contrast, incorporating legal statutes provides no additional benefit and may even degrade performance, as evidenced by the comparisons between S3 and S2, and S5 and S4, suggesting that naive use of legal texts can introduce noise. Finally, although S4 yields lower overall scores than S2, it more closely reflects real-world scenarios where flawed reasoning precedes incorrect decisions, making it a more challenging yet practically meaningful evaluation setting.

\subsection{Performance of SLMs (RQ3)}
Table~\ref{tab:slm} shows that on our \dataset, most existing state-of-the-art methods achieve higher accuracy but perform worse on other metrics. Compared with CAIL, \dataset places greater emphasis on error detection capability. Cases that are prone to misjudgment often involve low-frequency crimes with limited samples, which are inherently more difficult to learn than common charges. Therefore, Macro-F1 (MaF1) is a more meaningful evaluation metric in this setting. The relatively low MaF1 scores indicate both the challenging nature of our dataset and the limitations of current models in handling rare and error-prone cases.

\subsection{Performance Across Error Types (RQ4)}
The experimental results in Figure~\ref{fig:exp_error_types} show that the model performs relatively well in detecting charges and prison terms, while its performance on fine prediction is significantly weaker, especially under proportional fine errors (E6), indicating limitations in numerical understanding. Across different types of subsets, the overall classification performance remains low, highlighting the inherent difficulty of classifying correctly the fine-grained type of error. In correction tasks, the model achieves high performance in Erroneous Charge (E1) but struggles considerably with Omission of Charges (E2).  For prison terms and fines, classification metrics suggest similar difficulty levels for different error types (E3 \& E4) or even reverse difficulty levels (E5 \& E6), whereas regression metrics reveal that the model performs worse when dealing with more complex errors. These results suggest that continuous evaluation metrics provide a more comprehensive assessment of model capability, as interval-based classification may obscure meaningful differences—such as cases where interval boundaries overlap with true values or intervals are too broad to distinguish correct from incorrect predictions.

\begin{table}[!t]
\centering
\resizebox{\columnwidth}{!}{%
\begin{tabular}{lccc}
\toprule
Method & MaF1 (EDet) & MaF1 (EClass) & MaF1 (ECorr)\\
\midrule
Human & 49.50 & 100.00 & 70.10 \\
Qwen3-Max & 48.46 & 92.03 & 53.81 \\
\bottomrule
\end{tabular}
}
\caption{Human Performance on a Subset of \dataset for Error Detection (EDet), Error Classification (EClass), and Error Correction (ECorr).}
\label{tab:exp_human}
\end{table}

\subsection{Human Evaluation (RQ5)}
We report human performance on a randomly sampled subset of 150 samples in Table~\ref{tab:exp_human}. Annotators were provided with the same inputs as LLMs and evaluated using the same metrics. Human performance substantially exceeds that of current models, indicating that the task is well-defined and that there remains significant room for improvement.

\section{Conclusion}
Our work introduces \dataset, a large-scale resource designed to evaluate models' ability to perform \realtask by detecting, classifying and correcting judgment errors. Through comprehensive analysis, we reveal the limitations of current LLMs when classifying and correcting some obvious errors in judgments documents, exposing a gap between general language generation and reliable legal reasoning.
\dataset provides a new foundation for exploring judgment review tasks and improving legal reasoning in legal AI. We hope this resource catalyzes future efforts to build more capable legal AI models for \realtask.
\section*{Limitations}
Despite its contributions, this study has several limitations that should be acknowledged.
The scope of judgments is limited to Chinese court cases and does not include common law precedents from jurisdictions such as England and the United States. This trade-off was made to concentrate resources and ensure data quality during the initial research phase.
\dataset does not include legal interpretation or argumentation tasks, primarily because high-quality legal reasoning annotations heavily relying on experts are difficult to scale, and lack fine-grained domain evaluation criteria. Relying on existing large language models for assessment would introduce model bias, compromising objectivity.
We acknowledge these limitations and hope they will provide clear directions for future research.

\section*{Ethical Considerations}
Our work may involve the following ethical issues:
(1) Data Privacy and Offensive content. Our data is sourced from publicly available datasets, and sensitive information has been removed during the data construction process. We also use regular expressions to check whether sensitive information remains in the data and remove it. The case facts in the public judgment shall not contain any legally unfounded or unnecessarily offensive or inappropriate content, provided that they comply with legal provisions and judicial ethics. (2) Data Bias. Publicly available documents on the China Judgement Online exhibit selectivity and may vary in disclosure across different regions and court levels. This could introduce biases in the dataset regarding case types, geographic distribution, and temporal span, thereby affecting the benchmark's universality. (3) Legality and Compliance of Data Use. The purpose of constructing this dataset is to explore the limits of existing models in legal domain. Therefore, this dataset is intended solely for academic research and must not be used for commercial purposes or actual judicial decision-making. (4) Potential Impact on Judicial Fairness. AI-assisted judgment review may help standardize adjudication criteria, but it also risks overlooking the unique circumstances of each case. As a benchmark, \dataset does not provide any legal advice. Evaluation results from our datasets confirm that existing models still struggle to perform \realtask effectively. Given the complexity of legal cases, we emphasize that AI models should not replace human prosecutors or judges in the legal domain. We are committed to helping legal practitioners clarify the capabilities and boundaries of existing AI models, so they know when and how to use them.

Additionally, we used ChatGPT to polish the writing and are responsible for all the materials presented in this work.

\bibliography{custom}

@article{Zhang2023,
author = {Zhang, Han and Dou, Zhicheng and Zhu, Yutao and Wen, Ji-Rong},
title = {Contrastive Learning for Legal Judgment Prediction},
year = {2023},
issue_date = {October 2023},
publisher = {Association for Computing Machinery},
address = {New York, NY, USA},
volume = {41},
number = {4},
issn = {1046-8188},
url = {https://doi.org/10.1145/3580489},
doi = {10.1145/3580489},
journal = {ACM Trans. Inf. Syst.},
month = apr,
articleno = {113},
numpages = {25},
}

@article{Aletras2016,
  title={Predicting judicial decisions of the European Court of Human Rights: A natural language processing perspective},
  author={Aletras, Nikolaos and Tsarapatsanis, Dimitrios and Preo{\c{t}}iuc-Pietro, Daniel and Lampos, Vasileios},
  journal={PeerJ computer science},
  doi={10.7717/peerj-cs.93},
  volume={2},
  pages={e93},
  year={2016},
  publisher={PeerJ Inc.}
}

@inproceedings{Zhong2018,
    title = "Legal Judgment Prediction via Topological Learning",
    author = "Zhong, Haoxi  and
      Guo, Zhipeng  and
      Tu, Cunchao  and
      Xiao, Chaojun  and
      Liu, Zhiyuan  and
      Sun, Maosong",
    booktitle = "Proceedings of the 2018 Conference on Empirical Methods in Natural Language Processing",
    month = oct # "-" # nov,
    year = "2018",
    address = "Brussels, Belgium",
    publisher = "Association for Computational Linguistics",
    url = "https://aclanthology.org/D18-1390/",
    doi = "10.18653/v1/D18-1390",
    pages = "3540--3549",
}

@inproceedings{Chen2019,
    title = "Charge-Based Prison Term Prediction with Deep Gating Network",
    author = "Chen, Huajie  and Cai, Deng  and Dai, Wei  and Dai, Zehui  and Ding, Yadong",

    booktitle = "Proceedings of the 2019 Conference on Empirical Methods in Natural Language Processing and the 9th International Joint Conference on Natural Language Processing (EMNLP-IJCNLP)",
    month = nov,
    year = "2019",
    address = "Hong Kong, China",
    publisher = "Association for Computational Linguistics",
    url = "https://aclanthology.org/D19-1667/",
    doi = "10.18653/v1/D19-1667",
    pages = "6362--6367",
}

@inproceedings{Yue2021,
author = {Yue, Linan and Liu, Qi and Jin, Binbin and Wu, Han and Zhang, Kai and An, Yanqing and Cheng, Mingyue and Yin, Biao and Wu, Dayong},
title = {NeurJudge: A Circumstance-aware Neural Framework for Legal Judgment Prediction},
year = {2021},
isbn = {9781450380379},
publisher = {Association for Computing Machinery},
address = {New York, NY, USA},
url = {https://doi.org/10.1145/3404835.3462826},
doi = {10.1145/3404835.3462826},
booktitle = {Proceedings of the 44th International ACM SIGIR Conference on Research and Development in Information Retrieval},
pages = {973–982},
numpages = {10},
location = {Virtual Event, Canada},
series = {SIGIR '21}
}

@inproceedings{Feng2022,
    title = "Legal Judgment Prediction via Event Extraction with Constraints",
    author = "Feng, Yi  and Li, Chuanyi  and Ng, Vincent",
    booktitle = "Proceedings of the 60th Annual Meeting of the Association for Computational Linguistics (Volume 1: Long Papers)",
    month = may,
    year = "2022",
    address = "Dublin, Ireland",
    publisher = "Association for Computational Linguistics",
    url = "https://aclanthology.org/2022.acl-long.48/",
    doi = "10.18653/v1/2022.acl-long.48",
    pages = "648--664",
}

@misc{Cail2018,
      title={CAIL2018: A Large-Scale Legal Dataset for Judgment Prediction}, 
      author={Chaojun Xiao and Haoxi Zhong and Zhipeng Guo and Cunchao Tu and Zhiyuan Liu and Maosong Sun and Yansong Feng and Xianpei Han and Zhen Hu and Heng Wang and Jianfeng Xu},
      year={2018},
      eprint={1807.02478},
      archivePrefix={arXiv},
      primaryClass={cs.CL},
      url={https://arxiv.org/abs/1807.02478}, 
}

@inproceedings{SwissBenchmark2021,
    title = "{S}wiss-Judgment-Prediction: A Multilingual Legal Judgment Prediction Benchmark",
    author = {Niklaus, Joel  and  Chalkidis, Ilias  and  St{\"u}rmer, Matthias},
    booktitle = "Proceedings of the Natural Legal Language Processing Workshop 2021",
    month = nov,
    year = "2021",
    address = "Punta Cana, Dominican Republic",
    publisher = "Association for Computational Linguistics",
    url = "https://aclanthology.org/2021.nllp-1.3/",
    doi = "10.18653/v1/2021.nllp-1.3",
    pages = "19--35",
}

@inproceedings{Chalkidis2019,
    title = "Neural Legal Judgment Prediction in {E}nglish",
    author = "Chalkidis, Ilias  and Androutsopoulos, Ion  and Aletras, Nikolaos",
    booktitle = "Proceedings of the 57th Annual Meeting of the Association for Computational Linguistics",
    month = jul,
    year = "2019",
    address = "Florence, Italy",
    publisher = "Association for Computational Linguistics",
    url = "https://aclanthology.org/P19-1424/",
    doi = "10.18653/v1/P19-1424",
    pages = "4317--4323",
}

@inproceedings{LexEval2024,
 author = {Li, Haitao and Chen, You and Ai, Qingyao and Wu, Yueyue and Zhang, Ruizhe and Liu, Yiqun},
 booktitle = {Advances in Neural Information Processing Systems},
 doi = {10.52202/079017-0790},
 editor = {A. Globerson and L. Mackey and D. Belgrave and A. Fan and U. Paquet and J. Tomczak and C. Zhang},
 pages = {25061--25094},
 publisher = {Curran Associates, Inc.},
 title = {LexEval: A Comprehensive Chinese Legal Benchmark for Evaluating Large Language Models},
 url = {https://proceedings.neurips.cc/paper_files/paper/2024/file/2cb40fc022ca7bdc1a9a78b793661284-Paper-Datasets_and_Benchmarks_Track.pdf},
 volume = {37},
 year = {2024}
}

@inproceedings{Shui2023,
    title = "A Comprehensive Evaluation of Large Language Models on Legal Judgment Prediction",
    author = "Shui, Ruihao  and Cao, Yixin  and
      Wang, Xiang  and Chua, Tat-Seng",
    booktitle = "Findings of the Association for Computational Linguistics: EMNLP 2023",
    month = dec,
    year = "2023",
    address = "Singapore",
    publisher = "Association for Computational Linguistics",
    url = "https://aclanthology.org/2023.findings-emnlp.490/",
    doi = "10.18653/v1/2023.findings-emnlp.490",
    pages = "7337--7348",
}

@inproceedings{Yuan2024,
    title = "Can Large Language Models Grasp Legal Theories? Enhance Legal Reasoning with Insights from Multi-Agent Collaboration",
    author = "Yuan, Weikang  and Cao, Junjie  and Jiang, Zhuoren  and Kang, Yangyang  and Lin, Jun  and Song, Kaisong  and
      Lin, Tianqianjin  and Yan, Pengwei  and Sun, Changlong  and
      Liu, Xiaozhong",
    booktitle = "Findings of the Association for Computational Linguistics: EMNLP 2024",
    month = nov,
    year = "2024",
    address = "Miami, Florida, USA",
    publisher = "Association for Computational Linguistics",
    url = "https://aclanthology.org/2024.findings-emnlp.445/",
    doi = "10.18653/v1/2024.findings-emnlp.445",
    pages = "7577--7597",
}

@article{Hessick2008,
  title={Appellate Review of Sentencing Decisions},
  author={Hessick, Carissa Byrne and Hessick, F Andrew},
  journal={Ala. L. Rev.},
  volume={60},
  pages={1},
  year={2008},
  publisher={HeinOnline},
  url={https://heinonline.org/HOL/P?h=hein.journals/bamalr60&i=3}
}

@article{Orfield1936,
  title={History of Criminal Appeal in England},
  author={Orfield, Lester B},
  journal={Mo. L. Rev.},
  volume={1},
  pages={326},
  year={1936},
  publisher={HeinOnline},
  url={https://heinonline.org/HOL/P?h=hein.journals/molr1&i=341}
}

@article{Lindemann2020,
  title={Mechanisms for correcting judicial errors in Germany},
  author={Lindemann, Michael and Lienau, Fabienne},
  journal={Erasmus L. Rev.},
  volume={13},
  pages={73},
  year={2020},
  publisher={HeinOnline},
  url={https://heinonline.org/HOL/P?h=hein.journals/erasmus13&i=378}
}

@article{Hoquet1981,
  title={The French supreme court: The cour de cassation},
  author={Hoquet-McKee, Sophie},
  journal={City London L. Rev.},
  pages={1},
  year={1981},
  publisher={HeinOnline},
  url={https://heinonline.org/HOL/P?h=hein.journals/clonlr1981&i=8}
}

@article{Bai2006,
author = {Bai, Jianjun},
title = {An Empirical Study on the Change of Accusation},
journal = {Chinese Journal of Law},
number = {04},
pages = {51-62},
year = {2006},
issn = {1002-896X},
url = {https://faxueyanjiu.ajcass.com/Magazine/Show/?ID=92974}
}

@article{Woo1989,
  title={The right to a criminal appeal in the People's Republic of China},
  author={Woo, Margaret YK},
  journal={Yale J. Int'l L.},
  volume={14},
  pages={118},
  year={1989},
  publisher={HeinOnline},
  url={https://heinonline.org/HOL/P?h=hein.journals/yjil14&i=124}
}

@ARTICLE{Yue2024,
  author={Yue, Linan and Liu, Qi and Jin, Binbin and Wu, Han and An, Yanqing},
  journal={IEEE Transactions on Knowledge and Data Engineering}, 
  title={A Circumstance-Aware Neural Framework for Explainable Legal Judgment Prediction}, 
  year={2024},
  volume={36},
  number={11},
  pages={5453-5467},
  doi={10.1109/TKDE.2024.3387580}}

@inproceedings{Tyss2024,
    title = "Incorporating Precedents for Legal Judgement Prediction on {E}uropean Court of Human Rights Cases",
    author = "T.y.s.s, Santosh  and Elganayni, Mohamed Hesham  and S{\'o}jka, Stanis{\l}aw  and Grabmair, Matthias",
    booktitle = "Findings of the Association for Computational Linguistics: EMNLP 2024",
    month = nov,
    year = "2024",
    address = "Miami, Florida, USA",
    publisher = "Association for Computational Linguistics",
    url = "https://aclanthology.org/2024.findings-emnlp.214/",
    doi = "10.18653/v1/2024.findings-emnlp.214",
    pages = "3743--3750",
}

@inproceedings{Wu2023,
    title = "Precedent-Enhanced Legal Judgment Prediction with {LLM} and Domain-Model Collaboration",
    author = "Wu, Yiquan  and Zhou, Siying  and
      Liu, Yifei  and Lu, Weiming  and
      Liu, Xiaozhong  and Zhang, Yating  and
      Sun, Changlong  and Wu, Fei  and Kuang, Kun",
    booktitle = "Proceedings of the 2023 Conference on Empirical Methods in Natural Language Processing",
    month = dec,
    year = "2023",
    address = "Singapore",
    publisher = "Association for Computational Linguistics",
    url = "https://aclanthology.org/2023.emnlp-main.740/",
    doi = "10.18653/v1/2023.emnlp-main.740",
    pages = "12060--12075",
}

@inproceedings{Hu2018,
    title = "Few-Shot Charge Prediction with Discriminative Legal Attributes",
    author = "Hu, Zikun  and Li, Xiang  and Tu, Cunchao  and Liu, Zhiyuan  and Sun, Maosong",
    booktitle = "Proceedings of the 27th International Conference on Computational Linguistics",
    month = aug,
    year = "2018",
    address = "Santa Fe, New Mexico, USA",
    publisher = "Association for Computational Linguistics",
    url = "https://aclanthology.org/C18-1041/",
    pages = "487--498",
}

@inproceedings{Xu2020,
    title = "Distinguish Confusing Law Articles for Legal Judgment Prediction",
    author = "Xu, Nuo  and Wang, Pinghui  and Chen, Long  and
      Pan, Li  and Wang, Xiaoyan  and Zhao, Junzhou",
    booktitle = "Proceedings of the 58th Annual Meeting of the Association for Computational Linguistics",
    month = jul,
    year = "2020",
    address = "Online",
    publisher = "Association for Computational Linguistics",
    url = "https://aclanthology.org/2020.acl-main.280/",
    doi = "10.18653/v1/2020.acl-main.280",
    pages = "3086--3095",
}

@inproceedings{Gan2023,
    title = "Exploiting Contrastive Learning and Numerical Evidence for Confusing Legal Judgment Prediction",
    author = "Gan, Leilei  and Li, Baokui  and  Kuang, Kun  and
      Zhang, Yating  and  Wang, Lei  and Luu, Anh  and Yang, Yi  and
      Wu, Fei",
    booktitle = "Findings of the Association for Computational Linguistics: EMNLP 2023",
    month = dec,
    year = "2023",
    address = "Singapore",
    publisher = "Association for Computational Linguistics",
    url = "https://aclanthology.org/2023.findings-emnlp.814/",
    doi = "10.18653/v1/2023.findings-emnlp.814",
    pages = "12174--12185",
}

@inproceedings{Li2024,
    title = "From Graph to Word Bag: Introducing Domain Knowledge to Confusing Charge Prediction",
    author = "Li, Ang  and Chen, Qiangchao  and Wu, Yiquan  and Zhou, Xiang  and Kuang, Kun  and Wu, Fei  and Cai, Ming",
    booktitle = "Proceedings of the 2024 Joint International Conference on Computational Linguistics, Language Resources and Evaluation (LREC-COLING 2024)",
    month = may,
    year = "2024",
    address = "Torino, Italia",
    publisher = "ELRA and ICCL",
    url = "https://aclanthology.org/2024.lrec-main.659/",
    pages = "7469--7479",
}

@article{Bi2023,
	title = {Judicial knowledge-enhanced magnitude-aware reasoning for numerical legal judgment prediction},
	volume = {31},
	issn = {0924-8463, 1572-8382},
	url = {https://link.springer.com/10.1007/s10506-022-09337-4},
	doi = {10.1007/s10506-022-09337-4},
	language = {en},
	number = {4},
	urldate = {2025-11-27},
	journal = {Artificial Intelligence and Law},
	author = {Bi, Sheng and Zhou, Zhiyao and Pan, Lu and Qi, Guilin},
	month = dec,
	year = {2023},
	pages = {773--806},
}

@inproceedings{Guha2023,
 author = {Guha, Neel and Nyarko, Julian and Ho, Daniel and R\'{e}, Christopher and Chilton, Adam and K, Aditya and Chohlas-Wood, Alex and Peters, Austin and Waldon, Brandon and Rockmore, Daniel and Zambrano, Diego and Talisman, Dmitry and Hoque, Enam and Surani, Faiz and Fagan, Frank and Sarfaty, Galit and Dickinson, Gregory and Porat, Haggai and Hegland, Jason and Wu, Jessica and Nudell, Joe and Niklaus, Joel and Nay, John and Choi, Jonathan and Tobia, Kevin and Hagan, Margaret and Ma, Megan and Livermore, Michael and Rasumov-Rahe, Nikon and Holzenberger, Nils and Kolt, Noam and Henderson, Peter and Rehaag, Sean and Goel, Sharad and Gao, Shang and Williams, Spencer and Gandhi, Sunny and Zur, Tom and Iyer, Varun and Li, Zehua},
 booktitle = {Advances in Neural Information Processing Systems},
 editor = {A. Oh and T. Naumann and A. Globerson and K. Saenko and M. Hardt and S. Levine},
 pages = {44123--44279},
 publisher = {Curran Associates, Inc.},
 title = {LegalBench: A Collaboratively Built Benchmark for Measuring Legal Reasoning in Large Language Models},
 url = {https://proceedings.neurips.cc/paper_files/paper/2023/file/89e44582fd28ddfea1ea4dcb0ebbf4b0-Paper-Datasets_and_Benchmarks.pdf},
 volume = {36},
 year = {2023}
}

@inproceedings{Fei2024,
    title = "{L}aw{B}ench: Benchmarking Legal Knowledge of Large Language Models",
    author = "Fei, Zhiwei  and
      Shen, Xiaoyu  and
      Zhu, Dawei  and
      Zhou, Fengzhe  and
      Han, Zhuo  and
      Huang, Alan  and
      Zhang, Songyang  and
      Chen, Kai  and
      Yin, Zhixin  and
      Shen, Zongwen  and
      Ge, Jidong  and
      Ng, Vincent",
    booktitle = "Proceedings of the 2024 Conference on Empirical Methods in Natural Language Processing",
    month = nov,
    year = "2024",
    address = "Miami, Florida, USA",
    publisher = "Association for Computational Linguistics",
    url = "https://aclanthology.org/2024.emnlp-main.452/",
    doi = "10.18653/v1/2024.emnlp-main.452",
    pages = "7933--7962",
}

@inproceedings{LJPCheck2024,
    title = "{LJPC}heck: Functional Tests for Legal Judgment Prediction",
    author = "Zhang, Yuan  and
      Huang, Wanhong  and
      Feng, Yi  and
      Li, Chuanyi  and
      Fei, Zhiwei  and
      Ge, Jidong  and
      Luo, Bin  and
      Ng, Vincent",
    booktitle = "Findings of the Association for Computational Linguistics: ACL 2024",
    month = aug,
    year = "2024",
    address = "Bangkok, Thailand",
    publisher = "Association for Computational Linguistics",
    url = "https://aclanthology.org/2024.findings-acl.350/",
    doi = "10.18653/v1/2024.findings-acl.350",
    pages = "5878--5894",
}

@inproceedings{An2022,
    title = "Do Charge Prediction Models Learn Legal Theory?",
    author = "An, Zhenwei  and
      Huang, Quzhe  and
      Jiang, Cong  and
      Feng, Yansong  and
      Zhao, Dongyan",
    booktitle = "Findings of the Association for Computational Linguistics: EMNLP 2022",
    month = dec,
    year = "2022",
    address = "Abu Dhabi, United Arab Emirates",
    publisher = "Association for Computational Linguistics",
    url = "https://aclanthology.org/2022.findings-emnlp.275/",
    doi = "10.18653/v1/2022.findings-emnlp.275",
    pages = "3757--3768",
}

@inproceedings{Su2025,
author = {Su, Weihang and Yue, Baoqing and Ai, Qingyao and Hu, Yiran and Li, Jiaqi and Wang, Changyue and Zhang, Kaiyuan and Wu, Yueyue and Liu, Yiqun},
title = {JuDGE: Benchmarking Judgment Document Generation for Chinese Legal System},
year = {2025},
isbn = {9798400715921},
publisher = {Association for Computing Machinery},
address = {New York, NY, USA},
url = {https://doi.org/10.1145/3726302.3730295},
doi = {10.1145/3726302.3730295},
booktitle = {Proceedings of the 48th International ACM SIGIR Conference on Research and Development in Information Retrieval},
pages = {3573–3583},
numpages = {11},
location = {Padua, Italy},
series = {SIGIR '25}
}

@article{Wallace2005,
  title={Improving the Appellate Process Worldwide Through Maximizing Judicial Resources},
  author={Wallace, J Clifford},
  journal={Vand. J. Transnat'l L.},
  volume={38},
  pages={187},
  year={2005},
  publisher={HeinOnline},
  url={https://heinonline.org/HOL/P?h=hein.journals/vantl38&i=201}
}

@article{Lavie2016,
  title={Appellate courts and caseload pressure},
  author={Lavie, Shay},
  journal={Stan. L. \& Pol'y Rev.},
  volume={27},
  pages={57},
  year={2016},
  publisher={HeinOnline},
  url={https://heinonline.org/HOL/P?h=hein.journals/stanlp27&i=67}
}

@article{Tate1972,
  title={Containing the Law Explosion},
  author={Tate Jr, Albert},
  journal={Judicature},
  volume={56},
  pages={228},
  year={1972},
  publisher={HeinOnline},
  url={https://heinonline.org/HOL/P?h=hein.journals/judica56&i=230}
}

@InProceedings{Long2019,
author="Long, Shangbang and Tu, Cunchao and Liu, Zhiyuan and Sun, Maosong",
title="Automatic Judgment Prediction via Legal Reading Comprehension",
booktitle="Chinese Computational Linguistics",
year="2019",
publisher="Springer International Publishing",
address="Cham",
pages="558--572",
isbn="978-3-030-32381-3",
doi="10.1007/978-3-030-32381-3_45"
}

@inproceedings{Hwang2022,
 author = {Hwang, Wonseok and Lee, Dongjun and Cho, Kyoungyeon and Lee, Hanuhl and Seo, Minjoon},
 booktitle = {Advances in Neural Information Processing Systems},
 editor = {S. Koyejo and S. Mohamed and A. Agarwal and D. Belgrave and K. Cho and A. Oh},
 pages = {32537--32551},
 publisher = {Curran Associates, Inc.},
 title = {A Multi-Task Benchmark for Korean Legal Language Understanding and Judgement Prediction},
 url = {https://proceedings.neurips.cc/paper_files/paper/2022/file/d15abd14d5894eebd185b756541d420e-Paper-Datasets_and_Benchmarks.pdf},
 volume = {35},
 year = {2022}
}

@inproceedings{CaseSummarizer2016,
    title = "{C}ase{S}ummarizer: A System for Automated Summarization of Legal Texts",
    author = "Polsley, Seth  and
      Jhunjhunwala, Pooja  and
      Huang, Ruihong",
    editor = "Watanabe, Hideo",
    booktitle = "Proceedings of {COLING} 2016, the 26th International Conference on Computational Linguistics: System Demonstrations",
    month = dec,
    year = "2016",
    address = "Osaka, Japan",
    publisher = "The COLING 2016 Organizing Committee",
    url = "https://aclanthology.org/C16-2054/",
    pages = "258--262",
}

@article{Hu2024,
  title={Solicitation Behavior and Corruption Crimes},
  author={Hu, Changming},
  journal={Chengchi Law Review},
  volume={178},
  number={178},
  pages={1--35},
  year={2024},
  doi={10.53106/102398202024090178001},

}

@article{Zuo2018,
  title={Response to Litigation Explosion in China:Based on Empirical Analysis of the Trial Practice of the Court in W District in the Past Thirty Years},
  author={Zuo, Weimin},
  journal={Chinese Legal Science},
  year={2018},
  url={https://www.ncpssd.cn/Literature/articleinfo?id=675933296&synUpdateType=&type=journalArticle&from=Qikan_Article_Detail}
}

@article{Lawlor1963,
  title={What computers can do: analysis and prediction of judicial decisions},
  author={Lawlor, Reed C},
  journal={American Bar Association Journal},
  pages={337--344},
  year={1963},
  publisher={JSTOR},
  url={http://www.jstor.org/stable/25722338}
}

@article{Ulmer1963,
  title={Quantitative analysis of judicial processes: Some practical and theoretical applications},
  author={Ulmer, S Sidney},
  journal={Law and Contemporary Problems},
  volume={28},
  number={1},
  pages={164--184},
  year={1963},
  publisher={JSTOR},
  doi={10.2307/1190728}
}

@InProceedings{Liu2006,
author="Liu, Chao-Lin
and Hsieh, Chwen-Dar",
editor="Esposito, Floriana
and Ra{\'{s}}, Zbigniew W.
and Malerba, Donato
and Semeraro, Giovanni",
title="Exploring Phrase-Based Classification of Judicial Documents for Criminal Charges in Chinese",
booktitle="Foundations of Intelligent Systems",
year="2006",
publisher="Springer Berlin Heidelberg",
address="Berlin, Heidelberg",
pages="681--690",
isbn="978-3-540-45766-4",
doi={10.1007/11875604_75}
}

@inproceedings{Liu2022,
    title = "Augmenting Legal Judgment Prediction with Contrastive Case Relations",
    author = "Liu, Dugang  and
      Du, Weihao  and
      Li, Lei  and
      Pan, Weike  and
      Ming, Zhong",
    booktitle = "Proceedings of the 29th International Conference on Computational Linguistics",
    month = oct,
    year = "2022",
    address = "Gyeongju, Republic of Korea",
    publisher = "International Committee on Computational Linguistics",
    url = "https://aclanthology.org/2022.coling-1.235/",
    pages = "2658--2667",
}

@inproceedings{Shen2022,
 author = {Shen, Zejiang and Lo, Kyle and Yu, Lauren and Dahlberg, Nathan and Schlanger, Margo and Downey, Doug},
 title = {Multi-LexSum: Real-world Summaries of Civil Rights Lawsuits at Multiple Granularities},
 booktitle = {Advances in Neural Information Processing Systems},
 editor = {S. Koyejo and S. Mohamed and A. Agarwal and D. Belgrave and K. Cho and A. Oh},
 pages = {13158--13173},
 publisher = {Curran Associates, Inc.},

 url = {https://proceedings.neurips.cc/paper_files/paper/2022/file/552ef803bef9368c29e53c167de34b55-Paper-Datasets_and_Benchmarks.pdf},
 volume = {35},
 year = {2022}
}

@inproceedings{LegalAgentBench2025,
    title = "{L}egal{A}gent{B}ench: Evaluating {LLM} Agents in Legal Domain",
    author = "Li, Haitao  and
      Chen, Junjie  and
      Yang, Jingli  and
      Ai, Qingyao  and
      Jia, Wei  and
      Liu, Youfeng  and
      Lin, Kai  and
      Wu, Yueyue  and
      Yuan, Guozhi  and
      Hu, Yiran  and
      Wang, Wuyue  and
      Liu, Yiqun  and
      Huang, Minlie",
    booktitle = "Proceedings of the 63rd Annual Meeting of the Association for Computational Linguistics (Volume 1: Long Papers)",
    month = jul,
    year = "2025",
    address = "Vienna, Austria",
    publisher = "Association for Computational Linguistics",
    url = "https://aclanthology.org/2025.acl-long.116/",
    doi = "10.18653/v1/2025.acl-long.116",
    pages = "2322--2344",
    ISBN = "979-8-89176-251-0",
}

@inproceedings{Niklaus2025,
    title = "{S}wi{LT}ra-Bench: The {S}wiss Legal Translation Benchmark",
    author = {Niklaus, Joel  and
      Merane, Jakob  and
      Nenadic, Luka  and
      Ahmadi, Sina  and
      Gao, Yingqiang  and
      Chevalley, Cyrill A. H.  and
      Humbel, Claude  and
      G{\"o}sken, Christophe  and
      Tanzi, Lorenzo  and
      L{\"u}thi, Thomas  and
      Palombo, Stefan  and
      Poff, Spencer  and
      Yang, Boling  and
      Wu, Nan  and
      Guillod, Matthew  and
      Mami{\'e}, Robin  and
      Brunner, Daniel  and
      Pereyra, Julio  and
      Grupen, Niko},
    booktitle = "Proceedings of the 63rd Annual Meeting of the Association for Computational Linguistics (Volume 1: Long Papers)",
    month = jul,
    year = "2025",
    address = "Vienna, Austria",
    publisher = "Association for Computational Linguistics",
    url = "https://aclanthology.org/2025.acl-long.725/",
    doi = "10.18653/v1/2025.acl-long.725",
    pages = "14894--14916",
    ISBN = "979-8-89176-251-0",
}

@inproceedings{Rolshoven2025,
    title = "Unlocking Legal Knowledge: A Multilingual Dataset for Judicial Summarization in {S}witzerland",
    author = {Rolshoven, Luca  and
      Rasiah, Vishvaksenan  and
      Bose, Srinanda Br{\"u}gger  and
      Hostettler, Sarah  and
      Burkhalter, Lara  and
      St{\"u}rmer, Matthias  and
      Niklaus, Joel},
    booktitle = "Findings of the Association for Computational Linguistics: EMNLP 2025",
    month = nov,
    year = "2025",
    address = "Suzhou, China",
    publisher = "Association for Computational Linguistics",
    url = "https://aclanthology.org/2025.findings-emnlp.832/",
    doi = "10.18653/v1/2025.findings-emnlp.832",
    pages = "15382--15411",
    ISBN = "979-8-89176-335-7",
}

@inproceedings{Upadhya2025,
    title = "{L}ex{CL}i{PR}: Cross-Lingual Paragraph Retrieval from Legal Judgments",
    author = "Upadhya, Rohit  and
      T.y.s.s, Santosh",
    booktitle = "Proceedings of the 63rd Annual Meeting of the Association for Computational Linguistics (Volume 1: Long Papers)",
    month = jul,
    year = "2025",
    address = "Vienna, Austria",
    publisher = "Association for Computational Linguistics",
    url = "https://aclanthology.org/2025.acl-long.683/",
    doi = "10.18653/v1/2025.acl-long.683",
    pages = "13971--13993",
    ISBN = "979-8-89176-251-0",
}

@inproceedings{Seo2025,
    title = "{K}o{LEG}: On-the-Fly {K}orean Legal Knowledge Editing with Continuous Retrieval",
    author = "Seo, Jaehyung  and
      Jung, Dahyun  and
      Lee, Jaewook  and
      Chun, Yongchan  and
      Kim, Dongjun  and
      Ryu, Hwijung  and
      Shin, Donghoon  and
      Lim, Heuiseok",
    booktitle = "Findings of the Association for Computational Linguistics: EMNLP 2025",
    month = nov,
    year = "2025",
    address = "Suzhou, China",
    publisher = "Association for Computational Linguistics",
    url = "https://aclanthology.org/2025.findings-emnlp.489/",
    doi = "10.18653/v1/2025.findings-emnlp.489",
    pages = "9191--9217",
    ISBN = "979-8-89176-335-7",
}

@inproceedings{Yang2025,
    title = "{AD}-{LLM}: Benchmarking Large Language Models for Anomaly Detection",
    author = "Yang, Tiankai  and
      Nian, Yi  and
      Li, Li  and
      Xu, Ruiyao  and
      Li, Yuangang  and
      Li, Jiaqi  and
      Xiao, Zhuo  and
      Hu, Xiyang  and
      Rossi, Ryan A.  and
      Ding, Kaize  and
      Hu, Xia  and
      Zhao, Yue",
    booktitle = "Findings of the Association for Computational Linguistics: ACL 2025",
    month = jul,
    year = "2025",
    address = "Vienna, Austria",
    publisher = "Association for Computational Linguistics",
    url = "https://aclanthology.org/2025.findings-acl.79/",
    doi = "10.18653/v1/2025.findings-acl.79",
    pages = "1524--1547",
    ISBN = "979-8-89176-256-5"
}

@inproceedings{Felice2015,
    title = "Towards a standard evaluation method for grammatical error detection and correction",
    author = "Felice, Mariano  and
      Briscoe, Ted",
    booktitle = "Proceedings of the 2015 Conference of the North {A}merican Chapter of the Association for Computational Linguistics: Human Language Technologies",
    month = may # "–" # jun,
    year = "2015",
    address = "Denver, Colorado",
    publisher = "Association for Computational Linguistics",
    url = "https://aclanthology.org/N15-1060/",
    doi = "10.3115/v1/N15-1060",
    pages = "578--587"
}

@inproceedings{Schlichtkrull2023,
 author = {Schlichtkrull, Michael and Guo, Zhijiang and Vlachos, Andreas},
 booktitle = {Advances in Neural Information Processing Systems},
 editor = {A. Oh and T. Naumann and A. Globerson and K. Saenko and M. Hardt and S. Levine},
 pages = {65128--65167},
 publisher = {Curran Associates, Inc.},
 title = {AVeriTeC: A Dataset for Real-world Claim Verification with Evidence from the Web},
 url = {https://proceedings.neurips.cc/paper_files/paper/2023/file/cd86a30526cd1aff61d6f89f107634e4-Paper-Datasets_and_Benchmarks.pdf},
 volume = {36},
 year = {2023}
}

@inproceedings{FEVER2018,
    title = "{FEVER}: a Large-scale Dataset for Fact Extraction and {VER}ification",
    author = "Thorne, James  and
      Vlachos, Andreas  and
      Christodoulopoulos, Christos  and
      Mittal, Arpit",
    booktitle = "Proceedings of the 2018 Conference of the North {A}merican Chapter of the Association for Computational Linguistics: Human Language Technologies, Volume 1 (Long Papers)",
    month = jun,
    year = "2018",
    address = "New Orleans, Louisiana",
    publisher = "Association for Computational Linguistics",
    url = "https://aclanthology.org/N18-1074/",
    doi = "10.18653/v1/N18-1074",
    pages = "809--819",
}

@misc{GPT4o,
      title={GPT-4o System Card}, 
      author={Aaron Hurst and Adam Lerer and Adam P. Goucher and Adam Perelman and Aditya Ramesh and Aidan Clark and AJ Ostrow and Akila Welihinda and Alan Hayes and Alec Radford, et al.},
      year={2024},
      eprint={2410.21276},
      archivePrefix={arXiv},
      primaryClass={cs.CL},
      url={https://arxiv.org/abs/2410.21276}, 
}

@techreport{GPT51,
  author = {OpenAI},
  title = {GPT-5.1 Instant and GPT-5.1 Thinking System Card Addendum},
  institution = {OpenAI},
  year = {2025},
  month = {November},
  url = {https://cdn.openai.com/pdf/4173ec8d-1229-47db-96de-06d87147e07e/5_1_system_card.pdf},
  note = {Accessed: 2026-01-04}
}

@techreport{Minimax-m2,
  author = {MiniMax},
  title = {MiniMax M2 \& Agent: Ingenious in Simplicity},
  institution = {MiniMax},
  year = {2025},
  month = {October},
  url = {https://www.minimax.io/news/minimax-m2},
  note = {Accessed: 2026-01-04}
}

@misc{Qwen3,
      title={Qwen3 Technical Report}, 
      author={An Yang and Anfeng Li and Baosong Yang and Beichen Zhang and Binyuan Hui and Bo Zheng and Bowen Yu and Chang Gao and Chengen Huang and Chenxu Lv and Chujie Zheng and Dayiheng Liu and Fan Zhou and Fei Huang and Feng Hu and Hao Ge and Haoran Wei and Huan Lin and Jialong Tang and Jian Yang and Jianhong Tu and Jianwei Zhang and Jianxin Yang and Jiaxi Yang and Jing Zhou and Jingren Zhou and Junyang Lin and Kai Dang and Keqin Bao and Kexin Yang and Le Yu and Lianghao Deng and Mei Li and Mingfeng Xue and Mingze Li and Pei Zhang and Peng Wang and Qin Zhu and Rui Men and Ruize Gao and Shixuan Liu and Shuang Luo and Tianhao Li and Tianyi Tang and Wenbiao Yin and Xingzhang Ren and Xinyu Wang and Xinyu Zhang and Xuancheng Ren and Yang Fan and Yang Su and Yichang Zhang and Yinger Zhang and Yu Wan and Yuqiong Liu and Zekun Wang and Zeyu Cui and Zhenru Zhang and Zhipeng Zhou and Zihan Qiu},
      year={2025},
      eprint={2505.09388},
      archivePrefix={arXiv},
      primaryClass={cs.CL},
      url={https://arxiv.org/abs/2505.09388}, 
}

@misc{DeepSeekv32,
      title={DeepSeek-V3.2: Pushing the Frontier of Open Large Language Models}, 
      author={DeepSeek-AI and Aixin Liu and Aoxue Mei and Bangcai Lin and Bing Xue and Bingxuan Wang and Bingzheng Xu and Bochao Wu and Bowei Zhang and Chaofan Lin and Chen Dong and Chengda Lu and Chenggang Zhao and Chengqi Deng and Chenhao Xu and Chong Ruan and Damai Dai and Daya Guo and Dejian Yang and Deli Chen and Erhang Li and Fangqi Zhou and Fangyun Lin and Fucong Dai and Guangbo Hao and Guanting Chen and Guowei Li and H. Zhang and Hanwei Xu and Hao Li and Haofen Liang and Haoran Wei and Haowei Zhang and Haowen Luo and Haozhe Ji and Honghui Ding and Hongxuan Tang and Huanqi Cao and Huazuo Gao and Hui Qu and Hui Zeng and Jialiang Huang and Jiashi Li and Jiaxin Xu and Jiewen Hu and Jingchang Chen and Jingting Xiang and Jingyang Yuan and Jingyuan Cheng and Jinhua Zhu and Jun Ran and Junguang Jiang and Junjie Qiu and Junlong Li and Junxiao Song and Kai Dong and Kaige Gao and Kang Guan and Kexin Huang and Kexing Zhou and Kezhao Huang and Kuai Yu and Lean Wang and Lecong Zhang and Lei Wang and Liang Zhao and Liangsheng Yin and Lihua Guo and Lingxiao Luo and Linwang Ma and Litong Wang and Liyue Zhang and M. S. Di and M. Y Xu and Mingchuan Zhang and Minghua Zhang and Minghui Tang and Mingxu Zhou and Panpan Huang and Peixin Cong and Peiyi Wang and Qiancheng Wang and Qihao Zhu and Qingyang Li and Qinyu Chen and Qiushi Du and Ruiling Xu and Ruiqi Ge and Ruisong Zhang and Ruizhe Pan and Runji Wang and Runqiu Yin and Runxin Xu and Ruomeng Shen and Ruoyu Zhang and S. H. Liu and Shanghao Lu and Shangyan Zhou and Shanhuang Chen and Shaofei Cai and Shaoyuan Chen and Shengding Hu and Shengyu Liu and Shiqiang Hu and Shirong Ma and Shiyu Wang and Shuiping Yu and Shunfeng Zhou and Shuting Pan and Songyang Zhou and Tao Ni and Tao Yun and Tian Pei and Tian Ye and Tianyuan Yue and Wangding Zeng and Wen Liu and Wenfeng Liang and Wenjie Pang and Wenjing Luo and Wenjun Gao and Wentao Zhang and Xi Gao and Xiangwen Wang and Xiao Bi and Xiaodong Liu and Xiaohan Wang and Xiaokang Chen and Xiaokang Zhang and Xiaotao Nie and Xin Cheng and Xin Liu and Xin Xie and Xingchao Liu and Xingkai Yu and Xingyou Li and Xinyu Yang and Xinyuan Li and Xu Chen and Xuecheng Su and Xuehai Pan and Xuheng Lin and Xuwei Fu and Y. Q. Wang and Yang Zhang and Yanhong Xu and Yanru Ma and Yao Li and Yao Li and Yao Zhao and Yaofeng Sun and Yaohui Wang and Yi Qian and Yi Yu and Yichao Zhang and Yifan Ding and Yifan Shi and Yiliang Xiong and Ying He and Ying Zhou and Yinmin Zhong and Yishi Piao and Yisong Wang and Yixiao Chen and Yixuan Tan and Yixuan Wei and Yiyang Ma and Yiyuan Liu and Yonglun Yang and Yongqiang Guo and Yongtong Wu and Yu Wu and Yuan Cheng and Yuan Ou and Yuanfan Xu and Yuduan Wang and Yue Gong and Yuhan Wu and Yuheng Zou and Yukun Li and Yunfan Xiong and Yuxiang Luo and Yuxiang You and Yuxuan Liu and Yuyang Zhou and Z. F. Wu and Z. Z. Ren and Zehua Zhao and Zehui Ren and Zhangli Sha and Zhe Fu and Zhean Xu and Zhenda Xie and Zhengyan Zhang and Zhewen Hao and Zhibin Gou and Zhicheng Ma and Zhigang Yan and Zhihong Shao and Zhixian Huang and Zhiyu Wu and Zhuoshu Li and Zhuping Zhang and Zian Xu and Zihao Wang and Zihui Gu and Zijia Zhu and Zilin Li and Zipeng Zhang and Ziwei Xie and Ziyi Gao and Zizheng Pan and Zongqing Yao and Bei Feng and Hui Li and J. L. Cai and Jiaqi Ni and Lei Xu and Meng Li and Ning Tian and R. J. Chen and R. L. Jin and S. S. Li and Shuang Zhou and Tianyu Sun and X. Q. Li and Xiangyue Jin and Xiaojin Shen and Xiaosha Chen and Xinnan Song and Xinyi Zhou and Y. X. Zhu and Yanping Huang and Yaohui Li and Yi Zheng and Yuchen Zhu and Yunxian Ma and Zhen Huang and Zhipeng Xu and Zhongyu Zhang and Dongjie Ji and Jian Liang and Jianzhong Guo and Jin Chen and Leyi Xia and Miaojun Wang and Mingming Li and Peng Zhang and Ruyi Chen and Shangmian Sun and Shaoqing Wu and Shengfeng Ye and T. Wang and W. L. Xiao and Wei An and Xianzu Wang and Xiaowen Sun and Xiaoxiang Wang and Ying Tang and Yukun Zha and Zekai Zhang and Zhe Ju and Zhen Zhang and Zihua Qu},
      year={2025},
      eprint={2512.02556},
      archivePrefix={arXiv},
      primaryClass={cs.CL},
      url={https://arxiv.org/abs/2512.02556}, 
}

@misc{Glm2024,
      title={ChatGLM: A Family of Large Language Models from GLM-130B to GLM-4 All Tools},
      author={Team GLM and Aohan Zeng and Bin Xu and Bowen Wang and Chenhui Zhang and Da Yin and Diego Rojas and Guanyu Feng and Hanlin Zhao and Hanyu Lai and Hao Yu and Hongning Wang and Jiadai Sun and Jiajie Zhang and Jiale Cheng and Jiayi Gui and Jie Tang and Jing Zhang and Juanzi Li and Lei Zhao and Lindong Wu and Lucen Zhong and Mingdao Liu and Minlie Huang and Peng Zhang and Qinkai Zheng and Rui Lu and Shuaiqi Duan and Shudan Zhang and Shulin Cao and Shuxun Yang and Weng Lam Tam and Wenyi Zhao and Xiao Liu and Xiao Xia and Xiaohan Zhang and Xiaotao Gu and Xin Lv and Xinghan Liu and Xinyi Liu and Xinyue Yang and Xixuan Song and Xunkai Zhang and Yifan An and Yifan Xu and Yilin Niu and Yuantao Yang and Yueyan Li and Yushi Bai and Yuxiao Dong and Zehan Qi and Zhaoyu Wang and Zhen Yang and Zhengxiao Du and Zhenyu Hou and Zihan Wang},
      year={2024},
      eprint={2406.12793},
      archivePrefix={arXiv},
      primaryClass={id='cs.CL' full_name='Computation and Language' is_active=True alt_name='cmp-lg' in_archive='cs' is_general=False description='Covers natural language processing. Roughly includes material in ACM Subject Class I.2.7. Note that work on artificial languages (programming languages, logics, formal systems) that does not explicitly address natural-language issues broadly construed (natural-language processing, computational linguistics, speech, text retrieval, etc.) is not appropriate for this area.'}
}

@misc{KimiK2,
      title={Kimi K2: Open Agentic Intelligence}, 
      author={Kimi Team and Yifan Bai and Yiping Bao and Guanduo Chen and Jiahao Chen and Ningxin Chen and Ruijue Chen and Yanru Chen and Yuankun Chen and Yutian Chen and Zhuofu Chen and Jialei Cui and Hao Ding and Mengnan Dong and Angang Du and Chenzhuang Du and Dikang Du and Yulun Du and Yu Fan and Yichen Feng and Kelin Fu and Bofei Gao and Hongcheng Gao and Peizhong Gao and Tong Gao and Xinran Gu and Longyu Guan and Haiqing Guo and Jianhang Guo and Hao Hu and Xiaoru Hao and Tianhong He and Weiran He and Wenyang He and Chao Hong and Yangyang Hu and Zhenxing Hu and Weixiao Huang and Zhiqi Huang and Zihao Huang and Tao Jiang and Zhejun Jiang and Xinyi Jin and Yongsheng Kang and Guokun Lai and Cheng Li and Fang Li and Haoyang Li and Ming Li and Wentao Li and Yanhao Li and Yiwei Li and Zhaowei Li and Zheming Li and Hongzhan Lin and Xiaohan Lin and Zongyu Lin and Chengyin Liu and Chenyu Liu and Hongzhang Liu and Jingyuan Liu and Junqi Liu and Liang Liu and Shaowei Liu and T. Y. Liu and Tianwei Liu and Weizhou Liu and Yangyang Liu and Yibo Liu and Yiping Liu and Yue Liu and Zhengying Liu and Enzhe Lu and Lijun Lu and Shengling Ma and Xinyu Ma and Yingwei Ma and Shaoguang Mao and Jie Mei and Xin Men and Yibo Miao and Siyuan Pan and Yebo Peng and Ruoyu Qin and Bowen Qu and Zeyu Shang and Lidong Shi and Shengyuan Shi and Feifan Song and Jianlin Su and Zhengyuan Su and Xinjie Sun and Flood Sung and Heyi Tang and Jiawen Tao and Qifeng Teng and Chensi Wang and Dinglu Wang and Feng Wang and Haiming Wang and Jianzhou Wang and Jiaxing Wang and Jinhong Wang and Shengjie Wang and Shuyi Wang and Yao Wang and Yejie Wang and Yiqin Wang and Yuxin Wang and Yuzhi Wang and Zhaoji Wang and Zhengtao Wang and Zhexu Wang and Chu Wei and Qianqian Wei and Wenhao Wu and Xingzhe Wu and Yuxin Wu and Chenjun Xiao and Xiaotong Xie and Weimin Xiong and Boyu Xu and Jing Xu and Jinjing Xu and L. H. Xu and Lin Xu and Suting Xu and Weixin Xu and Xinran Xu and Yangchuan Xu and Ziyao Xu and Junjie Yan and Yuzi Yan and Xiaofei Yang and Ying Yang and Zhen Yang and Zhilin Yang and Zonghan Yang and Haotian Yao and Xingcheng Yao and Wenjie Ye and Zhuorui Ye and Bohong Yin and Longhui Yu and Enming Yuan and Hongbang Yuan and Mengjie Yuan and Haobing Zhan and Dehao Zhang and Hao Zhang and Wanlu Zhang and Xiaobin Zhang and Yangkun Zhang and Yizhi Zhang and Yongting Zhang and Yu Zhang and Yutao Zhang and Yutong Zhang and Zheng Zhang and Haotian Zhao and Yikai Zhao and Huabin Zheng and Shaojie Zheng and Jianren Zhou and Xinyu Zhou and Zaida Zhou and Zhen Zhu and Weiyu Zhuang and Xinxing Zu},
      year={2025},
      eprint={2507.20534},
      archivePrefix={arXiv},
      primaryClass={cs.LG},
      url={https://arxiv.org/abs/2507.20534}, 
}

@inproceedings{lawllm,
  title={LawLLM: Intelligent Legal System with Legal Reasoning and Verifiable Retrieval},
  author={Yue, Shengbin and Liu, Shujun and Zhou, Yuxuan and Shen, Chenchen and Wang, Siyuan and Xiao, Yao and Li, Bingxuan and Song, Yun and Shen, Xiaoyu and Chen, Wei and others},
  booktitle={International Conference on Database Systems for Advanced Applications},
  pages={304--321},
  year={2024},
  organization={Springer}
}

@misc{Farui,
  author = {{Alibaba Group}},
  title = {Tongyi Farui},
  year = {2026},
  url = {https://tongyi.aliyun.com/farui/home},
  note = {Accessed: 2026-01-04}
}

@misc{Gongbao:JudicialStatsChina,
  author       = {{PRC}},
  title        = {Judicial Statistics in {China}},
  year         = {2025},
  url          = {http://gongbao.court.gov.cn/ArticleList.html?serial_no=sftj},
  urldate      = {2025-09-19},
}

\appendix
\clearpage
\onecolumn

\section{Additional Details of \dataset}
\label{sec:appendix}

\subsection{License and Copyright Issue}
\label{app:license}
This dataset is derived and re-annotated from the original JuDGE\footnote{\url{https://github.com/oneal2000/JuDGE}}. The original work is dual-labeled with CC BY 4.0 in its publication~\citep{Su2025} and MIT License in its official GitHub repository, and we strictly comply with the MIT License (the definitive license for code/dataset distribution) for adaptation and re-annotation.
Modifications made to the original dataset include: (1) Re-annotating samples for \task; (2) Correcting erroneous labels in raw data; (3) Optimizing data format for model evaluation.
We have retained the original copyright notice and MIT License text in the documentation of our derived dataset, and no additional restrictions are imposed on the use, distribution, or adaptation of the derived data. 

\subsection{Anonymization}
Our dataset is derived from the original JudDGE~\citep{Su2025}, who claimed that "all dataset entries have undergone strict anonymization, eliminating
any personally identifiable data". Their raw data was obtained from China Judgments Online~\footnote{\url{https://wenshu.court.gov.cn/}}. According to the Provisions of the Supreme People’s Court on the Publication of Court Judgments on the Internet by People’s Courts~\footnote{\url{https://www.court.gov.cn/zixun/xiangqing/5867.html}}: 

\begin{figure*}[ht]
\centering
\begin{tcolorbox}[colback=white,colframe=black!75!white,colbacktitle=black!75!white,width=\textwidth,title=Article 7]

When publishing judgments and rulings on the Internet, the People’s Courts shall delete the following information:

\begin{enumerate}
    \item Personal information of natural persons, including but not limited to home addresses, contact details, identification card numbers, bank account numbers, and health information;
    \item Information relating to minors;
    \item Bank account numbers of legal persons and other organizations;
    \item Trade secrets;
    \item Other content that is not suitable for public disclosure.
\end{enumerate}

\end{tcolorbox}
\label{app:privacy}
\end{figure*}

Besides, we have removed or anonymized sensitive information from the dataset, such as home address and personally identifiable information. Therefore, using the dataset will not result in the disclosure of personal information.

\subsection{Data Annotation}
\label{app:annotation}
The dataset annotation involved three graduate students. They are all co-authors of this thesis, and no payment was provided. Before the annotation process, we invited a procurator of Procuratorate to explain the principles and common types of judgment errors, and the function of different components in the judgment documents. 
All annotators perform their annotations based on consistent annotation instructions.

\vspace{0.5 em}
\textbf{Construction of Erroneous Conviction (E1).} 
We first extract judgment documents pertaining to a single criminal charge, then extract sentencing factors from the judgments. Based on these factors, we construct a database of similar cases, categorizing cases with identical factors under the same charge. By matching similar cases with comparable factors among easily confused charges, we modify the original judgment accordingly. We replaced the reasoning in the original judgment documents with erroneous reasoning from the similar cases, and substituted the judgment with the incorrect offense, fine, and sentence from the similar cases. An example is shown in Table~\ref{case:E1_correction}.







\begin{table*}[t]
\centering
\resizebox{0.9\textwidth}{!}{
\begin{tabular}{p{2cm}|p{6cm}|p{6cm}}
 \toprule
Fact & \multicolumn{2}{p{12cm}}{Upon trial, it was ascertained that: 1. On April 27, 2008, the defendant XX, without obtaining the right to use three rooms of land at the East Sanjianfang base in Ying County, collected a deposit of RMB 8,600 from the victim Wang XX under the pretext of selling the three-room residential land to him. In the spring of 2010, when the victim Wang XX prepared to build a house on the land, he discovered that the land had already been built upon by others and was not owned by the defendant XX, which the defendant later admitted. 2. In early November 2009, Wang XX1 entrusted the defendant XX to use his residential land use certificate as collateral to borrow RMB 15,000 from others. On November 6 of the same year, the defendant XX used the land use certificate of the house as collateral to borrow RMB 60,000 from the victim Yuan XX, and handed RMB 15,000 to Wang XX1. On January 1, 2010, the defendant XX borrowed an additional RMB 10,000 from the victim Yuan XX, later changing his contact information. The victim Yuan XX filed a civil suit to claim rights against XX, which was tried in absentia, and in 2012 entered enforcement proceedings. 3. In 2009, Yang XX, through the defendant XX, borrowed RMB 70,000 from the victim Shi XX using a relocation housing contract, and later borrowed an additional RMB 40,000 from Shi XX. During this period, Yang XX repaid RMB 83,000 to the victim Shi XX, leaving RMB 27,000 used by the defendant XX. The defendant XX issued an IOU for the remaining RMB 27,000 and agreed to repay it in full by January 30, 2010. Later, the defendant XX left Ying County and changed his contact information.
 } \\
\hline
\multirow{2}{*}{Reasoning} 
& Original Judgement Document & Anomalous Judgment Document \\  \cline{2-3}
& Upon review, the Court finds that the defendant XX with the purpose of illegal possession, used methods of concealing the truth to defraud others of property, involving a huge amount, and that his conduct constitutes the crime of \textcolor{red}{contract fraud}. ... Therefore, in accordance with \textcolor{red}{Articles 244 and 64} of the Criminal Law of the People's Republic of China, the judgment is rendered as follows:
 & 
 Upon review, the Court finds that the defendant XX with the purpose of illegal possession, used methods of concealing the truth to defraud others of property, involving a huge amount, and that his conduct constitutes the crime of \textcolor{red}{fraud}. ... Therefore, in accordance with \textcolor{red}{Articles 25, 53, 64, 266, and 52} of the Criminal Law of the People's Republic of China, the judgment is rendered as follows:
 \\
\hline
\multirow{2}{*}{Judgment} 
& Original Judgement Document & Anomalous Judgment Document \\  \cline{2-3}
& I. The defendant XX is convicted of the crime of contract \textcolor{red}{fraud} and is sentenced to fixed-term imprisonment of \textcolor{red}{three years and six months}, and a fine of \textcolor{red}{RMB 5,000} ...
 & I. The defendant XX is convicted of the crime of \textcolor{red}{fraud} and is sentenced to fixed-term imprisonment of \textcolor{red}{three years and nine months}, and a fine of \textcolor{red}{RMB 80,000} ...
 \\
\hline
\multirow{2}{*}{Prison Term} 
& Original Judgement Document & Anomalous Judgment Document \\  \cline{2-3}
& fixed-term imprisonment of three years and six months
& fixed-term imprisonment of three years and nine months \\
\hline
\multirow{2}{*}{Fine} 
& Original Judgement Document & Anomalous Judgment Document \\  \cline{2-3}
& a fine of RMB 5000 & a fine of RMB 80000 \\
\hline
\multirow{2}{*}{Charge} 
& Original Judgement Document & Anomalous Judgment Document \\  \cline{2-3}
& contract fraud & fraud \\
\hline
\multirow{2}{*}{Law Articles} 
& Original Judgement Document & Anomalous Judgment Document \\  \cline{2-3}
& "244", "64" & "25", "53", "64", "266", "52" \\
\hline
Factor & \multicolumn{2}{p{12cm}}{\centering None}  \\
\bottomrule
\end{tabular}%
}
\caption{Example of Erroneous Conviction.}
\label{case:E1_correction}
\end{table*}

\vspace{0.5 em}
\textbf{Construction of Omission of Charges (E2).} 
We remove charges not on the target charge list from multiple charges, retaining only one charge, then extract sentencing factors from the judgment.
We match single-charge judgments with similar sentencing factors from our single-charge precedent database, replacing the original document's reasoning with erroneous precedent reasoning and substituting the judgment with the erroneous precedent's charge, fine, and sentence.
An example is shown in Table~\ref{case:E2_correction}.







\begin{table*}[t]
\centering
\resizebox{0.9\textwidth}{!}{
\begin{tabular}{p{2cm}|p{6cm}|p{6cm}}
 \toprule
Fact & \multicolumn{2}{p{12cm}}{Upon trial, it was ascertained that on March 13, 2018, at 23:00, the defendant XX, while driving a taxi under the influence of alcohol, drove in the opposite direction on the road in front of the north gate of Yiju Mediterranean in Kangping County and collided with a normally driven car with license plate number Zhe E8T792. Traffic police officers XX and XX of the Kangping County Public Security Bureau arrived at the scene to handle the accident. When the defendant XX was brought to the outpatient department of Kangping County People's Hospital for blood collection, he kicked Officer XX with his feet. After forced blood collection, he punched XX in the face and verbally abused him, and was later subdued. The accident was caused by XX’s drunk driving, and he was found fully responsible for the accident. Forensic examination detected ethanol in the defendant XX’s venous blood at a concentration of 231.7 mg/100 ml.
 } \\
\hline
\multirow{2}{*}{Reasoning} 
& Original Judgement Document & Anomalous Judgment Document \\  \cline{2-3}
& The Court holds that the defendant... committed the crimes of \textcolor{red}{obstruction of official}duties and \textcolor{red}{dangerous driving}. The facts of the crimes are clear, and the evidence is reliable and sufficient. The defendant shall be sentenced for obstruction of official duties and dangerous driving, to be combined for multiple crimes. In accordance with \textcolor{red}{Articles 277, 133-1, 65(1), 67(3), 42, 44, 45, 47, 69(2), and 52} of the Criminal Law of the People's Republic of China, the judgment is rendered as follows:
& The Court holds that the defendant's conduct constitutes the crime of \textcolor{red}{dangerous driving}. The public prosecution's allegations are upheld. In accordance with \textcolor{red}{Articles 133-1(1), 67(1), 42, 44, 52, 77(1), 69, 45, and 47} of the Criminal Law of the People's Republic of China, the judgment is rendered as follows:
 
 \\
\hline
\multirow{2}{*}{Judgment} 
& Original Judgement Document & Anomalous Judgment Document \\  \cline{2-3}
& The defendant XX is convicted of the crime of \textcolor{red}{obstruction of official duties} and is sentenced to fixed-term imprisonment of ten months; the defendant XX is convicted of the crime of \textcolor{red}{dangerous driving} and is sentenced to criminal detention of three months and fined \textcolor{red}{RMB 3,000}. The sentences are combined, and the final execution is fixed-term imprisonment of \textcolor{red}{ten months} and a fine of RMB 3,000 (already paid) ...
 & The defendant is convicted of the crime of \textcolor{red}{dangerous driving} and sentenced to criminal detention of two months and fined \textcolor{red}{RMB 2,000}, which, after revocation of probation and combination of sentences with fixed-term imprisonment of \textcolor{red}{eleven months}, results in the execution of fixed-term imprisonment of eleven months and a fine of RMB 2,000 ...
 \\
\hline
\multirow{2}{*}{Prison Term} 
& Original Judgement Document & Anomalous Judgment Document \\  \cline{2-3}
& fixed-term imprisonment of ten months
& fixed-term imprisonment of eleven months\\
\hline
\multirow{2}{*}{Fine} 
& Original Judgement Document & Anomalous Judgment Document \\  \cline{2-3}
& a fine of RMB 3,000 & a fine of RMB 2,000  \\
\hline
\multirow{2}{*}{Charge} 
& Original Judgement Document & Anomalous Judgment Document \\  \cline{2-3}
& "obstruction of official duties", "dangerous driving" & "dangerous driving" \\
\hline
\multirow{2}{*}{Law Articles} 
& Original Judgement Document & Anomalous Judgment Document \\  \cline{2-3}
& "69",
        "67",
        "44",
        "47",
        "42",
        "133",
        "45",
        "277",
        "52" &  "69",
            "67",
            "47",
            "77",
            "133",
            "42",
            "45",
            "44",
            "52" \\
\hline
Factor & \multicolumn{2}{p{12cm}}{\centering All}  \\
\bottomrule
\end{tabular}%
}
\caption{Example of Omission of Charges.}
\label{case:E2_correction}
\end{table*}

\vspace{0.5 em}
\textbf{Construction of Sentencing Outside Limits (E3).} 
For convictions involving monetary amounts, we extract the amount from the judgment and determine the sentencing tier based on that amount. For convictions not involving monetary amounts, we extract the description of the tier from the judgment document. For judgments without sentencing factors: If the sentence falls within a single tier, we adjust the sentence to fall outside that tier. If the sentence spans multiple tiers, we adjust the sentence to fall within the adjacent tier.
An example is shown in Table~\ref{case:E3_correction}.






\begin{table*}[t]
\centering
\resizebox{0.9\textwidth}{!}{
\begin{tabular}{p{2cm}|p{6cm}|p{6cm}}
 \toprule
Fact & \multicolumn{2}{p{12cm}}{Upon trial, it was ascertained that the defendant XX, for the purpose of illegal profit and at the instigation of Qu Mou and Yu Mou (both handled in separate cases), issued false Shandong Province ordinary value-added tax (VAT) invoices on multiple occasions between June 22, 2018 and February 26, 2019. In the absence of any genuine business transactions between herself and enterprises including Tongxiang Zhongyan Fur Co., Ltd., Binhai Haotai Garment Co., Ltd., and Dingzhou Lu Mou Garment, she went to the Muping District State Taxation Bureau of Yantai City and the First Taxation Branch of the State Taxation Administration of Yantai City Muping District to falsely issue a total of 27 Shandong Province ordinary VAT invoices. Through Wang Mou, these invoices were sold to the aforementioned enterprises, involving a total invoice amount of RMB 15,638,295. The defendant XX illegally obtained approximately RMB 10,000, which was used for her personal and family daily expenses. After the case was discovered, the defendant’s relatives returned the illicit proceeds on her behalf in the amount of RMB 10,000 and voluntarily paid the fine.
 } \\
\hline
Reasoning & \multicolumn{2}{p{12cm}}{Upon review, the Court finds that the defendant XX’s conduct of illegally selling Shandong Province ordinary value-added tax invoices that could be used to fraudulently obtain export tax refunds or to offset tax liabilities constitutes the crime of illegally selling invoices used for fraudulently obtaining export tax refunds or for offsetting tax liabilities. The crime charged by the prosecuting authority is established. The defendant XX pleaded guilty and accepted punishment in court, and her relatives returned the illicit proceeds on her behalf and voluntarily paid the fine, which constitutes mitigating circumstances warranting leniency. The defense counsel’s opinion requesting a mitigated punishment for the defendant is well-founded and is adopted. The sentencing recommendation of the prosecuting authority is appropriate. In accordance with Paragraphs 1 and 3 of Article 209, and Articles 64, 52, and 53 of the Criminal Law of the People's Republic of China, the judgment is rendered as follows:
 } \\
\hline
\multirow{2}{*}{Judgment} 
& Original Judgement Document & Anomalous Judgment Document \\  \cline{2-3}
& I. The defendant XX ... is sentenced to fixed-term imprisonment of \textcolor{red}{one year} ...
 & I. The defendant XX ... is sentenced to fixed-term imprisonment of \textcolor{red}{six years and two months} ...
 \\
\hline
\multirow{2}{*}{Prison Term} 
& Original Judgement Document & Anomalous Judgment Document \\  \cline{2-3}
& fixed-term imprisonment of one year
& fixed-term imprisonment of six years and two months \\
\hline
Law Articles & \multicolumn{2}{p{12cm}}{\centering "53",
        "64",
        "209",
        "52"
 } \\
\hline
Factor & \multicolumn{2}{p{12cm}}{\centering None}  \\
\bottomrule
\end{tabular}%
}
\caption{Example of Sentencing Outside Limits.}
\label{case:E3_correction}
\end{table*}

\vspace{0.5 em}
\textbf{Construction of Failure to Consider Sentencing Factors (E4).} If the case involves mitigating circumstances and the actual sentence falls within the first 50\% of the sentencing range, we will adjust the erroneous sentence to a value between 0.9 times the maximum sentence and the maximum sentence itself.
If the case involves aggravating circumstances and the actual sentence falls within the latter 50\% of the sentencing range, we will adjust the erroneous sentence to a value between the minimum sentence and 1.1 times the minimum sentence.
We will also remove the description of the relevant circumstances under “reasoning” and the corresponding citation numbers for the applicable legal provisions.
An example is shown in Table~\ref{case:E4_correction}.






\begin{table*}[t]
\centering
\resizebox{0.9\textwidth}{!}{
\begin{tabular}{p{2cm}|p{6cm}|p{6cm}}
 \toprule
Fact & \multicolumn{2}{p{12cm}}{The Yongxin County People's Procuratorate alleges that on June 2, 2015, Yin Mou (already convicted) registered and established a certain guarantee company and, without obtaining the qualification to raise funds from the public, engaged in the illegal absorption of public deposits. At Yin Mou’s request, the defendant XX provided his personal Rural Commercial Bank account (Account No.: 62×××86) and the corresponding password to Yin Mou for the purpose of raising funds from the public. Between September 10, 2015 and May 7, 2016, Yin Mou used the bank account provided by the defendant XX to illegally absorb public deposits from the public and transferred funds totaling RMB 3.76 million. The defendant XX voluntarily surrendered himself, which constitutes a circumstance of voluntary surrender and, in accordance with the law, may warrant a mitigated or reduced punishment; as a first-time offender, he may also be given a discretionary mitigated punishment. In light of the foregoing, it is recommended that the defendant XX be sentenced to fixed-term imprisonment of less than one year or criminal detention, and be fined. The defendant XX raises no objection to the criminal facts and charge alleged by the prosecuting authority, pleaded guilty and accepted punishment in court, and requested leniency and the application of probation.
 } \\
\hline
\multirow{2}{*}{Reasoning} 
& Original Judgement Document & Anomalous Judgment Document \\  \cline{2-3}
& Upon review, the Court finds that the defendant XX ... after committing the crime, \textcolor{red}{voluntarily surrendered himself and truthfully confessed to his criminal conduct, constituting voluntary surrender, and may therefore be given a mitigated punishment in accordance with the law}; he pleaded guilty and accepted punishment in court and is a first-time offender, and may therefore be given a discretionary mitigated punishment. ... Accordingly, in accordance with Item (1) of \textcolor{red}{Paragraph 1 of Article 191, Paragraph 1 of Article 67, Articles 72, Paragraphs 2 and 3 of Article 73, and Article 52} of the Criminal Law of the People's Republic of China, the judgment is rendered as follows:
 & 
 Upon review, the Court finds that the defendant XX ... his criminal conduct, \textcolor{red}{in accordance with the law}; he pleaded guilty and accepted punishment in court and is a first-time offender, in accordance with the law. Accordingly, in accordance with Item (1) of \textcolor{red}{Paragraph 1 of Article 191, Articles 72, Paragraphs 2 and 3 of Article 73, and Article 52} of the Criminal Law of the People's Republic of China, the judgment is rendered as follows:
 \\
\hline
\multirow{2}{*}{Judgment} 
& Original Judgement Document & Anomalous Judgment Document \\  \cline{2-3}
& fixed-term imprisonment of six months
 & fixed-term imprisonment of four years and seven months
 \\
\hline
\multirow{2}{*}{Prison Term} 
& Original Judgement Document & Anomalous Judgment Document \\  \cline{2-3}
& The defendant XX is convicted of the crime of money laundering and is sentenced to fixed-term imprisonment of six months ...
 & The defendant XX is convicted of the crime of money laundering and is sentenced to fixed-term imprisonment of four years and seven months ... \\
\hline
\multirow{2}{*}{Law Articles} 
& Original Judgement Document & Anomalous Judgment Document \\  \cline{2-3}
&"73",
        "67",
        "72",
        "191",
        "52"  
        & "73",
        "72",
        "191",
        "52"  \\
\hline
Factor & \multicolumn{2}{p{12cm}}{\centering Mitigating Factor}  \\
\bottomrule
\end{tabular}%
}
\caption{Example of Failure to Consider Sentencing Factors.}
\label{case:E4_correction}
\end{table*}

\vspace{0.5 em}
\textbf{Construction of fixed-amount Fine Outside Limits (E5).} We extract descriptions of the tiers from documents for offenses without specified monetary thresholds. The erroneous construction of fixed-amount sentencing involves modifying fines beyond the prescribed range.
An example is shown in Table~\ref{case:E5_correction}.






\begin{table*}[t]
\centering
\resizebox{0.9\textwidth}{!}{
\begin{tabular}{p{2cm}|p{6cm}|p{6cm}}
 \toprule
Fact & \multicolumn{2}{p{12cm}}{The prosecuting authority alleges that, beginning in May 2013, the defendant Qiu, for the purpose of obtaining illegal profits, sold various types of illegally manufactured invoices. On August 1, public security officers apprehended Qiu at the residence he rented, located at Room 102, No. 5, Jixiang Lane, Jingxi Street, Baiyun District of this city, and seized 6,536 illegally manufactured invoices (as identified upon appraisal, including 3,284 local tax invoices and 108 special value-added tax invoices, all of which were counterfeit), one batch of invoice stubs and express delivery stubs, as well as tools used for counterfeiting, including a computer and a printer.
 } \\
\hline
\multirow{2}{*}{Reasoning} 
& Original Judgement Document & Anomalous Judgment Document \\  \cline{2-3}
& Upon review, the Court finds that the defendant Qiu ... After committing the crime, the defendant Qiu  \textcolor{red}{truthfully confessed to his criminal conduct and may therefore be given a mitigated punishment}. ... In accordance with  \textcolor{red}{Paragraph 2 of Article 209, Paragraph 3 of Article 67, Articles 52, 53, and 64} of the Criminal Law of the People's Republic of China, the judgment is rendered as follows:
 & Upon review, the Court finds that the defendant Qiu ...  \textcolor{red}{the defendant Qiu his criminal conduct}, ... In accordance with  \textcolor{red}{Paragraph 2 of Article 209, Articles 52, 53, and 64} of the Criminal Law of the People's Republic of China, the judgment is rendered as follows:
 
 \\
\hline
\multirow{2}{*}{Judgment} 
& Original Judgement Document & Anomalous Judgment Document \\  \cline{2-3}
& I. The defendant Qiu ... and is sentenced to a fine of  \textcolor{red}{RMB 50,000} ...
 & I. The defendant Qiu ... and is sentenced to a fine of  \textcolor{red}{RMB 9,077} ...
 \\
\hline
\multirow{2}{*}{Fine} 
& Original Judgement Document & Anomalous Judgment Document \\  \cline{2-3}
& a fine of RMB 50,000 & a fine of RMB 9,077   \\
\hline
\multirow{2}{*}{Law Articles} 
& Original Judgement Document & Anomalous Judgment Document \\  \cline{2-3}
&  "53",
        "64",
        "67",
        "209",
        "52"
        & 
        "53",
        "64",
        "209",
        "52"\\
\hline
Factor & \multicolumn{2}{p{12cm}}{\centering Mitigating Factor}  \\
\bottomrule
\end{tabular}%
}
\caption{Example of Fixed-amount Fine Outside Limits.}
\label{case:E5_correction}
\end{table*}

\vspace{0.5 em}
\textbf{Construction of Percentage-based Fine Outside Limits (E6).} We extract the monetary amounts from sentencing documents for offenses involving specified monetary thresholds, determine the sentencing tier based on these amounts, and then calculate the corresponding fine range according to the Criminal Law based on the extracted fines. 
An example is shown in Table~\ref{case:E6_correction}.






\begin{table*}[t]
\centering
\resizebox{0.9\textwidth}{!}{
\begin{tabular}{p{2cm}|p{6cm}|p{6cm}}
 \toprule
Fact & \multicolumn{2}{p{12cm}}{The Huangmei County People's Procuratorate alleges that during the year 2012, the defendant XX, who at the time was responsible for credit operations at the Huangjinshan Branch of Hubei Bank, became acquainted through work with Lin Mou (already adjudicated). In order to fulfill deposit-collection targets, XX required Lin Mou to invest RMB 50 million by depositing the funds at the Huangjinshan Branch of Hubei Bank for a term of five years, with interest calculated at a monthly rate of 5.5\%. XX further promised that when Lin Mou required liquidity, she would assist him in obtaining six-month loans by pledging the certificates of deposit as collateral. Thereafter, Lin Mou reached an agreement with the person in charge of the Huangjinshan Branch of Hubei Bank under which the monthly loan interest rate ranged from 5.04\% to 5.47\%. ... From lending out the RMB 20 million obtained through this bank loan, Lin Mou collected RMB 100,000 in interest from Liu Mou and RMB 200,000 in interest from Liu Mou, and additionally allowed XX to obtain RMB 80,000 in benefits from Liu Mou. However, in the joint crime of high-interest relending, the defendant XX played an auxiliary role and was an accessory, to whom the relevant provisions apply. The defendant XX’s illegal gains are to be handled in accordance with the applicable provisions.
 } \\
\hline
\multirow{2}{*}{Reasoning} 
& Original Judgement Document & Anomalous Judgment Document \\  \cline{2-3}
& Upon review, the Court finds that the defendant XX colluded with others ... and should be given a mitigated punishment. After the crime was discovered, the defendant XX \textcolor{red}{truthfully confessed to her criminal conduct, constituting a confession}, and may therefore be given a mitigated punishment in accordance with the law. ... In accordance with \textcolor{red}{Article 175, Article 27, Paragraph 3 of Article 67, Articles 72, 73, 64, and 61} of the Criminal Law of the People's Republic of China, the judgment is rendered as follows:
 & Upon review, the Court finds that the defendant XX colluded with others ... and should. After the crime was discovered, the defendant XX \textcolor{red}{her criminal conduct} ... In accordance with \textcolor{red}{Article 175, Article 27, Articles 72, 73, 64, and 61} of the Criminal Law of the People's Republic of China, the judgment is rendered as follows:
 
 \\
\hline
\multirow{2}{*}{Judgment} 
& Original Judgement Document & Anomalous Judgment Document \\  \cline{2-3}
& I. The defendant XX ... and is sentenced to a fine of \textcolor{red}{RMB 160,000} ...
 & I. The defendant XX ... and is sentenced to a fine of \textcolor{red}{RMB 432,171} ...
 \\
\hline
\multirow{2}{*}{Fine} 
& Original Judgement Document & Anomalous Judgment Document \\  \cline{2-3}
& a fine of RMB 160,000 & a fine of RMB 432,171   \\
\hline
\multirow{2}{*}{Law Articles} 
& Original Judgement Document & Anomalous Judgment Document \\  \cline{2-3}
&   "73",
        "64",
        "67",
        "175",
        "72",
        "61",
        "27"
        & 
        "73",
        "64",
        "175",
        "72",
        "61",
        "27"\\
\hline
Factor & \multicolumn{2}{p{12cm}}{\centering Mitigating Factor}  \\
\bottomrule
\end{tabular}%
}
\caption{Example of Percentage-based Fine Outside Limits.}
\label{case:E6_correction}
\end{table*}

\section{Additional Details on Experiments}
\subsection{Hyperparameter settings}

\begin{table*}[t] 
    \centering
    \begin{minipage}{0.48\textwidth}
        \centering
\resizebox{0.96\columnwidth}{!}{
\begin{tabular}{lccc}
\toprule
Hyperparameter & G-LLMs & R-LLMs & D-LLMs \\
\midrule
frequency\_penalty & 0 & 0 & 0 \\
logprobs & false & false & false  \\
presence\_penalty & 0 & 0 & 0  \\
temperature & 0 & 0 & 0 \\
max\_output\_tokens & 2048 & 2048 & 1024 \\

\bottomrule
\end{tabular}
}
\caption{Hyperparameter settings. G-LLMs is short for General-Domain LLMs,  R-LLMs is short for Reasoning-Enhanced LLMs, and D-LLMs is short for Domain-Specific LLMs.}
\label{tab:app_hyper}
    \end{minipage}
    \hfill 
    \begin{minipage}{0.48\textwidth}
        \centering
\resizebox{0.9\columnwidth}{!}{
\begin{tabular}{lcc}
\toprule
Model & Training Set & GPU hours  \\
\midrule
LADAN+MIL & CAIL small &  1 \\
TopJudge  & CAIL small & 0.6  \\
NeurJudge & CAIL small &  4 \\
LADAN+MIL & CAIL big &  29 \\
TopJudge  & CAIL big & 17 \\
NeurJudge & CAIL big & 98 \\
\bottomrule
\end{tabular}
}
\caption{Computational Budget}
\label{tab:app_gpu}
    \end{minipage}
\end{table*}

We report the hyperparameters used for the LLMs in Table~\ref{tab:app_hyper}. When evaluating LLMs, we set the temperature to 0 to minimize the variance introduced by random sampling. We employed the same hyperparameter settings as in the original papers when training SLMs on CAIL2018 dataset. 

\subsection{Computational Budget}
All experiments were conducted on an Ubuntu server equipped with a 32-core Intel(R) Xeon(R) Silver 4110 CPU @ 2.10GHz and 1 Tesla V100-PCIE-32GB. Additionally, we fine-tuned legal judgment prediction models to evaluate their performance. The CUDA version was 13.0, the Python version was 3.1, the PyTorch version was 2.4.0+cu121, and the tensorflow-gpu version was 1.14.0. The GPU hours of training are shown in Table~\ref{tab:app_gpu}.

\subsection{Instruction Following Rate}
\label{app:instruction_following_rate}
\begin{table}[!t]
\centering
\begin{tabular}{lccc}
\toprule
Models & Error Detection & Error Classification & Error Correction \\
\midrule
GPT-5.1  & 100.00 & 98.13 & 99.98 \\
Qwen3-Max & 100.00 & 100.00 & 100.00 \\
DeepSeek-V3.2-Chat & 99.95 & 99.92 & 99.97 \\
GLM-4.6-Chat & 99.97 & 87.57 & 98.81 \\
MiniMax-M2  & 100.00 & 100.00 & 100.00\\
Qwen3-14B-Chat & 100.00 & 100.00 & 100.00 \\
Qwen3-8B-Chat & 100.00 & 99.84 & 99.57\\
GPT-4o  & 100.00 & 99.79 & 100.00 \\
Qwen3-Max-preview-Thinking  & 99.70 & 99.70 & 99.97 \\
DeepSeek-V3.2-Thinking & 99.85 & 96.44 & 99.65 \\
Kimi-K2-Thinking & 99.34 & 98.61 & 99.73\\
Qwen3-Next-80B-A3B-Thinking & 99.94  & 99.91 & 99.94\\
Farui-Plus  & 99.85 & 99.76 & 99.91 \\
LawLLM-7B  & 100.00 & 100.00 & 100.00 \\

\bottomrule
\end{tabular}
\caption{Instruction Following Rate (\%) on three sub-tasks of \dataset.}
\label{tab:instruction_following}
\end{table}
Table~\ref{tab:instruction_following} shows how well the models follow the prompt instructions. If a model’s output can be correctly parsed by the evaluation script, we consider the model to be able to follow the instructions of \task. As can be seen, the instruction-following rate for the classification task is lower than that of the other two tasks. This is because many models tend to under-predict errors, which cannot be revealed by accuracy alone. Therefore, we did not include this portion of the data in our calculations. As a result, the performance of the error classification task is relatively higher than the models' real performance.

\subsection{Prompts}
The prompt templates for the three tasks are presented below. Specifically, for the error correction task, we adjust the prompts based on the three target variables: charge, prison term, and fine.

\begin{figure*}[ht]
\centering
\begin{tcolorbox}[colback=white,colframe=black!75!white,colbacktitle=black!75!white,width=\textwidth,title=\pmtDetection]
\# Role and Mission

You are a professional criminal legal analysis expert specializing in determining whether a judgment contains errors based on case facts (Fact), the reasoning process and legal provisions cited in the original judgment (Reasoning), and the judgment itself (Judgment).

\#\# Core Task

Based on the user-provided “Case Facts (Fact)”, “Original Judgment's Reasoning Process and Legal Basis (Reasoning)”, and “Original Judgment (Judgment)”, accomplish one thing:

1. Determine whether the original judgment contains errors (output only true/false);

\#\# Analysis Basis

1. Strictly adhere to relevant provisions of China's Criminal Law, integrating core elements from the case facts (e.g., criminal acts, involved amounts, circumstances of the crime, restitution status, etc.);

2. Focus on examining whether the reasoning process and legal provisions cited in the original judgment (Reasoning) are logically coherent, legally accurate, and form a complete closed loop with the case facts;

3. Error criteria: Explicit conflict between the original judgment/reasoning/cited legal provisions and case facts/legal statutes, excluding reasonable sentencing range variations;

4. Focus solely on the three core elements—“Charge, Prison Term, and Fine”—excluding other procedural content in the judgment (e.g., appeal deadlines, restitution methods).

\#\# Output Format (Strictly required; otherwise deemed invalid)
Output only a JSON string containing one field: llm\_iserror (bool). Example:

\begin{verbatim}
{
    “llm_iserror”: true
}
\end{verbatim}
or
\begin{verbatim}
{
    “llm_iserror”: false
}
\end{verbatim}
\end{tcolorbox}
\caption{\pmtDetection}
\label{prompt:prompt_for_error_detection}
\end{figure*}

\begin{figure*}[ht]
\centering
\begin{tcolorbox}[colback=white,colframe=black!75!white,colbacktitle=black!75!white,width=\textwidth,title=\pmtClassification]
\# Role and Mission

You are a professional criminal legal analysis expert, skilled at identifying error types by analyzing case facts (Fact), the reasoning process of the original judgment (Reasoning), and the original judgment (Judgment).

\#\# Core Task

Based on the user-provided “Case Facts (Fact)”, “Original Judgment Reasoning Process (Reasoning)”, and “Original Judgment (Judgment)”, accomplish one thing:

1. Identify the error type (select only from the following 3 categories; no other types):

   - Charge Error: The charge determined in the judgment does not align with the legal provisions corresponding to the case facts;
   
   - Prison Term Error: The prison term is correctly identified, but the imposed sentence duration (e.g., term of detention, fixed-term imprisonment) does not comply with legal provisions or sentencing circumstances;
   
   - Fine Error: The charge and prison term are correctly identified, but the imposed fine amount does not comply with legal provisions or sentencing circumstances;

\#\# Basis for Analysis

1. Strictly adhere to relevant provisions of China's Criminal Law, integrating core elements of the case facts (e.g., criminal conduct, involved amount, circumstances of the crime, restitution status, etc.);

2. Focus on reviewing whether the reasoning process in the original judgment is logically coherent, whether the application of law is accurate, and whether it forms a complete closed loop with the case facts;

3. Error criteria: The original judgment or reasoning process conflicts explicitly with case facts or legal provisions, rather than representing reasonable sentencing range variations;

4. Focus solely on the three core elements—“Charge, Prison Term, and Fine”—excluding other procedural content in the judgment (e.g., appeal deadlines, restitution methods).

\#\# Output Format (Strict compliance is mandatory; otherwise, the output is invalid)

Output only a JSON-formatted string containing one field: 

\begin{verbatim}
{
    “llm_error_type”: “Charge Error”
}
\end{verbatim}
\end{tcolorbox}
\caption{\pmtClassification}
\label{prompt:prompt_for_error_classification}
\end{figure*}

\begin{figure*}[ht]
\centering
\begin{tcolorbox}[colback=white,colframe=black!75!white,colbacktitle=black!75!white,width=\textwidth,title=\pmtChargeCorrection]
\# Role and Mission

You are a professional criminal legal analysis expert, skilled at determining the correct answer based on case facts (Fact), the reasoning process of the original judgment (Reasoning), and the original judgment (Judgment).

\#\# Core Task

Based on the user-provided “Case Facts (Fact)”, “Original Judgment Reasoning Process (Reasoning)”, and “Original Judgment (Judgment)”, accomplish one thing:

1. The known judgment contains a charge error:. You must provide the correct charge. Output only the correct charge (e.g., “Fraud”) without additional explanation.

\#\# Analytical Basis

1. Strictly adhere to relevant provisions of China's Criminal Law, integrating core elements from the case facts (e.g., criminal acts, involved amounts, circumstances of the crime, restitution status, etc.);

2. Focus on examining whether the original judgment's reasoning process is logically coherent, whether the application of law is accurate, and whether it forms a complete closed loop with the case facts;

3. Criteria for erroneous judgments: The original judgment or reasoning process conflicts explicitly with the case facts or legal provisions, rather than representing reasonable variations in sentencing ranges;

4. Focus solely on the three core elements—“Charge, Prison Term, and Fine”—excluding other procedural content in the judgment (e.g., appeal deadlines, restitution methods).

\#\# Output Format (Strict compliance is mandatory; otherwise, the output is invalid)

Output only a JSON-formatted string containing one field: llm\_right (string). Do not provide any explanation, analysis, reasoning, or commentary. Example:

\begin{verbatim}
{
    “llm_right”: “Fraud”
}
\end{verbatim}

\end{tcolorbox}
\caption{\pmtChargeCorrection}
\label{prompt:prompt_for_charge_error_correction}
\end{figure*}

\begin{figure*}[ht]
\centering
\begin{tcolorbox}[colback=white,colframe=black!75!white,colbacktitle=black!75!white,width=\textwidth,title=\pmtTermCorrection]
\# Role and Mission

You are a professional criminal legal analysis expert, skilled at determining the correct answer based on case facts (Fact), the reasoning process of the original judgment (Reasoning), and the original judgment (Judgment).

\#\# Core Task

Based on the user-provided “Case Facts (Fact)”, “Original Judgment Reasoning (Reasoning)”, and “Original Judgment (Judgment)”, accomplish one thing:

1.The known judgment contains a prison term error:. You must provide the correct prison term. Output only the correct prison term (e.g., “5 months of criminal detention”) without additional explanation.

\#\# Analytical Basis

1. Strictly adhere to relevant provisions of China's Criminal Law, integrating core elements from the case facts (e.g., criminal acts, involved amounts, circumstances of the crime, restitution status, etc.);

2. Focus on examining whether the original judgment's reasoning process is logically coherent, whether the application of law is accurate, and whether it forms a complete closed loop with the case facts;

3. Criteria for erroneous judgments: The original judgment or reasoning process conflicts explicitly with case facts or legal provisions, rather than representing reasonable variations in sentencing ranges;

4. Focus solely on the three core elements—“Charge, Prison Term, and Fine”—excluding other procedural content in the judgment (e.g., appeal deadlines, restitution methods).

\#\# Output Format (Strict compliance is mandatory; otherwise, the output is invalid)

Output only a JSON-formatted string containing one field: llm\_right (string). Do not include any explanations, analysis, reasoning, or annotations. Example:

\begin{verbatim}
{
    “llm_right”: “fixed-term imprisonment of one year and 5 months”
}
\end{verbatim}

\end{tcolorbox}
\caption{\pmtTermCorrection}
\label{prompt:prompt_for_term_error_correction}
\end{figure*}

\begin{figure*}[ht]
\centering
\begin{tcolorbox}[colback=white,colframe=black!75!white,colbacktitle=black!75!white,width=\textwidth,title=\pmtFineCorrection]
\# Role and Mission
You are a professional criminal legal analysis expert, skilled at determining the correct answer based on case facts (Fact), the reasoning process of the original judgment (Reasoning), and the original judgment (Judgment).

\#\# Core Task

Based on the user-provided “Case Facts (Fact)”, “Original Judgment Reasoning (Reasoning)”, and “Original Judgment (Judgment)”, accomplish one thing:

1. The known judgment contains a fine error. You must provide the correct fine amount. Output only the correct fine amount (e.g., ‘a fine of RMB 2000’) without any additional explanation.

\#\# Analytical Basis

1. Strictly adhere to relevant provisions of China's Criminal Law, integrating core elements from the case facts (e.g., criminal acts, involved amounts, circumstances of the crime, restitution status, etc.);

2. Focus on reviewing whether the original judgment's reasoning process is logically coherent, whether the application of law is accurate, and whether it forms a complete closed loop with the case facts;

3. Criteria for erroneous judgment: The original judgment or reasoning process conflicts explicitly with the case facts or legal provisions, rather than representing a reasonable variation in sentencing range;

4. Focus solely on the three core elements—“Charge, Prison Term, and Fine”—excluding other procedural content in the judgment (e.g., appeal deadlines, restitution methods).

\#\# Output Format (Strict compliance is mandatory; otherwise, the output is invalid)

Output only a JSON-formatted string containing one field: llm\_right (string). Do not include any explanations, analysis, reasoning, or annotations. Example:

\begin{verbatim}
{
    “llm_right”: “a fine of RMB 2000”
}
\end{verbatim}

\end{tcolorbox}
\caption{\pmtFineCorrection}
\label{prompt:prompt_for_fine_error_correction}
\end{figure*}

\end{document}